\documentclass[lettersize,journal]{IEEEtran}
\usepackage{amsmath,amsfonts}
\usepackage{algorithmic}
\usepackage{array}
\usepackage[caption=false,font=normalsize,labelfont=sf,textfont=sf]{subfig}
\usepackage{textcomp}
\usepackage{stfloats}
\usepackage{url}
\usepackage{verbatim}
\usepackage{graphicx}
\usepackage{svg}
\usepackage{multirow}

\usepackage{siunitx} 
\usepackage{threeparttable}
\def\BibTeX{{\rm B\kern-.05em{\sc i\kern-.025em b}\kern-.08em
    T\kern-.1667em\lower.7ex\hbox{E}\kern-.125emX}}
\usepackage{balance}

\usepackage{booktabs}
\begin{document}
\title{ColorVein: Colorful Cancelable Vein Biometrics}
	\author{Yifan Wang,~\IEEEmembership{Student Member,~IEEE,} Jie Gui,~\IEEEmembership{Senior Member,~IEEE,} Xinli Shi,~\IEEEmembership{Senior Member,~IEEE,} Linqing Gui, Yuan Yan Tang,~\IEEEmembership{Life Fellow,~IEEE,} and James Tin-Yau Kwok,~\IEEEmembership{Fellow,~IEEE}
	\thanks{This work was supported in part by the grant of the National Science Foundation of China under Grant 62172090; Start-up Research Fund of Southeast University under Grant RF1028623097. We thank the Big Data Computing Center of Southeast University for providing the facility support on the numerical calculations. (Corresponding author: Jie Gui.)}
	\thanks{Y. Wang, J. Gui and X. Shi are with the School of Cyber Science and Engineering, Southeast
		University, Nanjing 210000, China. J. Gui is also with Engineering Research Center of Blockchain Application, Supervision and Management (Southeast University), Ministry of Education; Purple Mountain Laboratories, Nanjing 210000, China (e-mail: 230239767@seu.edu.cn, guijie@seu.edu.cn).}
	\thanks{L. Gui is with the College of Computer, Nanjing University of Posts
		and Telecommunications, Nanjing 210003, China (e-mail: guilq@njupt.edu.cn).}
	\thanks{Y.Tang is with the Department of Computer and Information Science, University of Macao, Macao 999078, China (e-mail: yytang@um.edu.mo).}
	\thanks{J. Kwok is with the Department of Computer Science and Engineering, The Hong Kong University of Science and Technology, Hong
		Kong 999077, China (e-mail: jamesk@cse.ust.hk).}
	\thanks{This paper has been accepted by IEEE Transactions on Information Forensics and Security}
}

\markboth{Journal of \LaTeX\ Class Files,~Vol.~18, No.~9, September~2020}%
{How to Use the IEEEtran \LaTeX \ Templates}

\maketitle

\begin{abstract}
		Vein recognition technologies have become one of the primary solutions for high-security identification systems. However, the issue of biometric information leakage can still pose a serious threat to user privacy and anonymity.
		Currently, there is no cancelable biometric template generation scheme specifically designed for vein biometrics. Therefore, this paper proposes an innovative cancelable vein biometric generation scheme: ColorVein. Unlike previous cancelable template generation schemes, ColorVein does not destroy the original biometric features and introduces additional color information to grayscale vein images. This method significantly enhances the information density of vein images by transforming static grayscale information into dynamically controllable color representations through interactive colorization. ColorVein allows users/administrators to define a controllable pseudo-random color space for grayscale vein images by editing the position, number, and color of hint points, thereby generating protected cancelable templates. Additionally, we propose a new secure center loss to optimize the training process of the protected feature extraction model, effectively increasing the feature distance between enrolled users and any potential impostors. Finally, we evaluate ColorVein's performance on all types of vein biometrics, including recognition performance, unlinkability, irreversibility, and revocability, and conduct security and privacy analyses. ColorVein achieves competitive performance compared with state-of-the-art methods.
\end{abstract}


\section{Introduction}
\label{sec:intro}
\IEEEPARstart{V}{ein} biometrics is an identification technique that uses intrinsic physiological features within the human body.
When near infrared (NIR) light illuminates human tissues, the hemoglobin in the vein blood vessels absorbs more NIR light than the surrounding tissues, resulting in a distinctive shadow during the imaging process, and presenting the vein pattern in the image \cite{hou2022finger}.
These images are acquired as infrared images (grayscale), which contain feature information such as pattern, brightness, and contrast. These features are extracted and used for subsequent processing and analysis, forming the baseline for identity matching. Cancelable biometrics (CBs) is a method that enhances system security by applying intentional and repeatable distortion transformations to original biometric features. It allows biometric templates to be reissued and revoked repeatedly, similar to traditional keys, and enables authentication in the encrypted domain, thereby achieving both repeatability and revocability of biometric traits. This technology employs specific algorithms to apply controllable transformations to the original features, effectively preventing the theft or reconstruction of biometric information and providing stronger security and privacy protection for biometric systems~\cite{10620353}.

Computer vision technology has become an important tool in the field of biometric recognition, often inspired by the human visual system, which serves as a solid foundation for its understanding and development~\cite{wang2023residual}. Color images are found all around us.
The human visual system has the innate ability to recognize thousands of colors, but is relatively limited in its ability to discriminate at the grayscale level \cite{buhrmester2019evaluating}.
Color images can carry richer information than grayscale images.
Similar to common computer vision tasks, color is also critical in biometrics \cite{yip2002contribution}, e.g., in face \cite{xin2023large}, iris \cite{wei2024multi}, and gait \cite{huang2024integral} recognition.
However, vein recognition has a unique challenge: unlike face, iris, or gait, where infrared grayscale images \cite{he2019adversarial} \cite{luo2021partial} or natural color images \cite{huang2024integral} can be selectively acquired, regular vein features can only be captured as infrared grayscale images due to their inherent imaging principles \cite{zhao2024vpcformer}.

\begin{figure}[t]
	\centering
	\includegraphics[scale=1,width=0.45\textwidth]{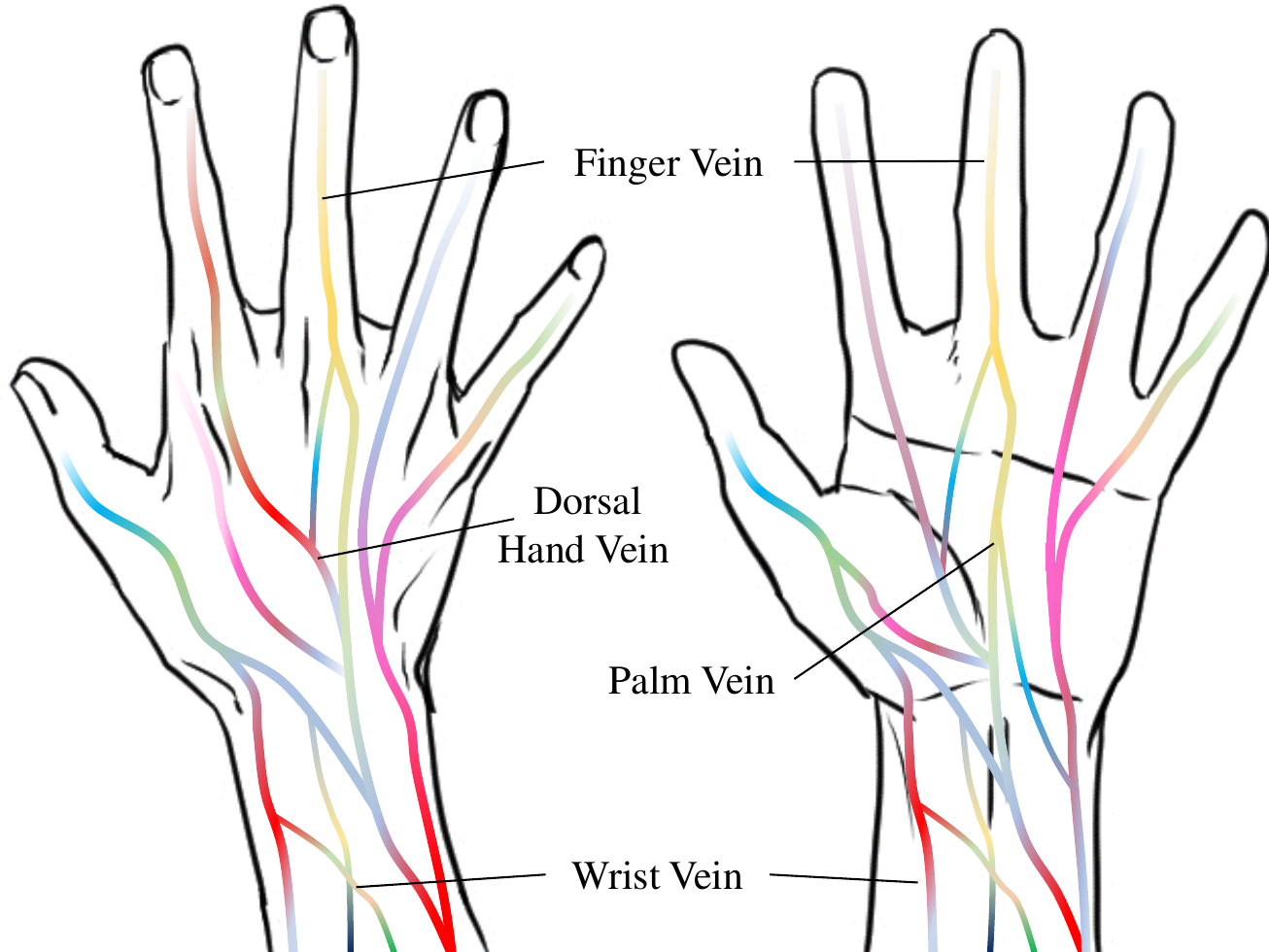}
	\caption{Vein biometrics: finger vein, palm vein, dorsal hand vein, wrist vein.}
	\label{fig1}
\end{figure}

Vein recognition is a typical fine-grained classification task \cite{huang2021joint, song2022eifnet}. Grayscale image only provides limited feature information, this results in the differences among vein images mainly in pattern features, and such differences are susceptible to be further attenuated by other unstable factors in the image (e.g., translation, rotation, or contrast change, etc.) \cite{10620353}. Inspired by color biometrics and image colorization techniques, we would like to try to colorize vein images to inject stable and information-rich color information. We to significantly enhance the vein image information density, expanding the original single grayscale channel into a multi-dimensional RGB three-channel representation, expanding the information space available for feature extraction, and helping the feature extractor or authentication model to capture finer inter-class differences, improving the biometric system performance and robustness.

The color space is editable, and this provide a new concept for designing cancelable vein biometrics using colorization. Specifically, it can be achieved through interactive colorization schemes.
According to ISO/IEC 24745 standard on biometric information protection \cite{iso2022}, cancelable biometrics is defined as performing intentional and repeatable distortion transformation operations on biometric signals to achieve repeated release and revocability of biometrics \cite{ratha2001enhancing}.
By introducing an interactive colorization scheme, we can transform static grayscale information into dynamically controllable colored representations.
It allows the user or system administrator to define a controlled pseudo-random color space for grayscale vein images, by editing the location, number and color of hint points.
This interactive colorization scheme answers three key questions, “Where to be colorized?” , i.e., control the position of the hint points, enable selective colorization on specific areas of the vein image.
“What color?” by customizing the color of each hint point to create a unique color mapping scheme.
And “How rich is the color?” , i.e., adjust the number and distribution of hint points to control the complexity and diversity of the colorization results.
It can fit into the core requirements of CB systems. By simply changing the color mapping scheme, it can easily generate new cancelable templates, enabling revocability.
There is no special mapping relationship between the colored vein used for matching and the original biometric, so the original biometric cannot be reconstructed from the protected vector, ensuring unlinkability.
Different applications can use different color mapping schemes to prevent cross-database linking, and unlinkability can be satisfied.
Finally the suitable color mapping can preserve or even enhance the recognition performance.

Therefore, we propose a novel cancelable vein biometric generation scheme, The main contributions of this paper are as follows:

	\begin{itemize}
	\item ColorVein is an interactive infrared vein image colorization scheme that transforms static grayscale information into dynamic and controllable colored representations.
	It allows users/administrators to define a controlled pseudo-random color space for grayscale vein images by editing the location, number and color of hint points to generate protected cancelable templates. 
	
	\item We propose a novel objective function: the secure center (SC) loss function, which is used in the supervised training process of the protection feature extraction model. 
	By constructing four sample pairs, SC loss can effectively increase the feature distance between legitimate enrolled users and various potential impostors. Ensure that illegal users in cross-application or impostors with stolen tokens cannot access the system, to enhance the security and reliability.
	
	\item We evaluate ColorVein scheme's recognition performance, unlinkability, irreversibility and revocability on all vein biometrics (finger, palm, dorsal hand and wrist). Furthermore we analyzed the security and privacy of ColorVein. We also compare ColorVien with representative cancelable biometric generation schemes.
	
	\end{itemize}

The rest of the paper is organized as follows. Section~\ref{sec:related} briefly the related works. Section~\ref{sec:ColorVein} introduces the proposed ColorVein scheme. Section~\ref{sec:expt} presents the experimental setup, results, and analysis. Section~\ref{s&p} conducts the privacy, security, and CB attributes analysis. Finally, the paper concludes with Section~\ref{sec:conclusion}.
\section{Related work}
\label{sec:related}
In this section, we briefly review related work on cancelable vein biometrics and image colorization.
\subsection{Cancelable Vein Biometrics}
Cancelable transformation is a way for biometric template protection. It transforms the original biometric template $T$ into a different domain using a token and a transformation function $F$, thereby generating a protected template $T_P$. In identification systems, only the transformed template is used. If the template is compromised, it is only necessary to change the token to regenerate the protected template \cite{ahmad2019lightweight}. Despite the fact that there has been a lot of research on cancelable schemes for biometrics such as fingerprint, face, iris and palmprints~\cite{yang2024physics}, there is still a lack of research on cancelable schemes for vein biometrics. Fortunately, many cancelable biometrics schemes proposed for other biometrics can be transferred and applied to vein biometrics. The most representative methods include block remapping \cite{ratha2001enhancing}, mesh warping \cite{10620353}, Biohashing \cite{jin2004biohashing} and Bloom filter \cite{bloom} \cite{rathgeb2014application}.

In the field of cancelable biometrics, many approaches have been proposed for various biometric modalities. Among the representative methods mentioned above, block remapping \cite{ratha2001enhancing}, mesh warping~\cite{10620353} and Bloom filters~\cite{bloom} were initially used for iris biometrics, while biohashing~\cite{jin2004biohashing} was first proposed for fingerprint recognition. Other methods such as geometric transformations~\cite{ratha2007generating}, random projections~\cite{wu2023multi} and minutiae transformations~\cite{lee2007alignment} have also been extensively studied in fingerprint recognition. In face biometrics, various improved hashing algorithms~\cite{gao2024protected} have been developed to generate cancelable templates. In addition, several deep learning based methods~\cite{wang2024make} have been devised to create cancelable face features.

Piciucco et al. \cite{piciucco2016cancelable} discussed the possibility of applying block remapping and mesh warping to design cancelable vein biometrics. Kauba et al. \cite{kauba2022towards} further investigate and compare the performance and template protection attributes of block remapping, mesh warping and Bloom filters on vein biometrics. Nayar et al. \cite{nayar2021graph} then proposed a graph-based scheme for cancelable vein biometrics generation. Meanwhile, hash-based methods have been widely studied and discussed in vein biometrics. This includes modified biohashing \cite{10620353, shahreza2021towards, yang2019securing, zhong2018palm}, as well as modified schemes based on locally sensitive hashing \cite{kirchgasser2019finger, kirchgasser2020finger,choudhary2022multi}. These feature transformation operations often compromise the biometric information used for identification, such as vein branches, junction points, and branch angles. By adding "pseudo-random information" to the template to achieve revocability, this may result in a significant decrease in identification performance.

In addition, some researchers introduced cryptographic techniques to design cancelable biometric features. For example, Ren et al. \cite{ren2021finger} encrypted vein images using RSA (Rivest-Shamir-Adleman) encryption and used the encrypted images to train deep learning models for recognition. Aherrahrou et al. \cite{aherrahrou2023novel} used local dissimilarity maps to project features in order to design cancelable vein templates. Yang et al. \cite{yang2018securing} used fuzzy commitment scheme (FCS) to encrypt biometric features and healthcare data to design cancelable vein identity. The cancelable attributes produced by the above methods are derived from the specific cryptographic techniques used and are not biometric related. 

Although all of the above methods have been transferred or applied on vein biometrics, there is still a lack of a cancelable biometric generation scheme constructed specifically for vein characteristics.

\begin{figure*}[t]
	\includegraphics[scale=1,width=0.99\textwidth]{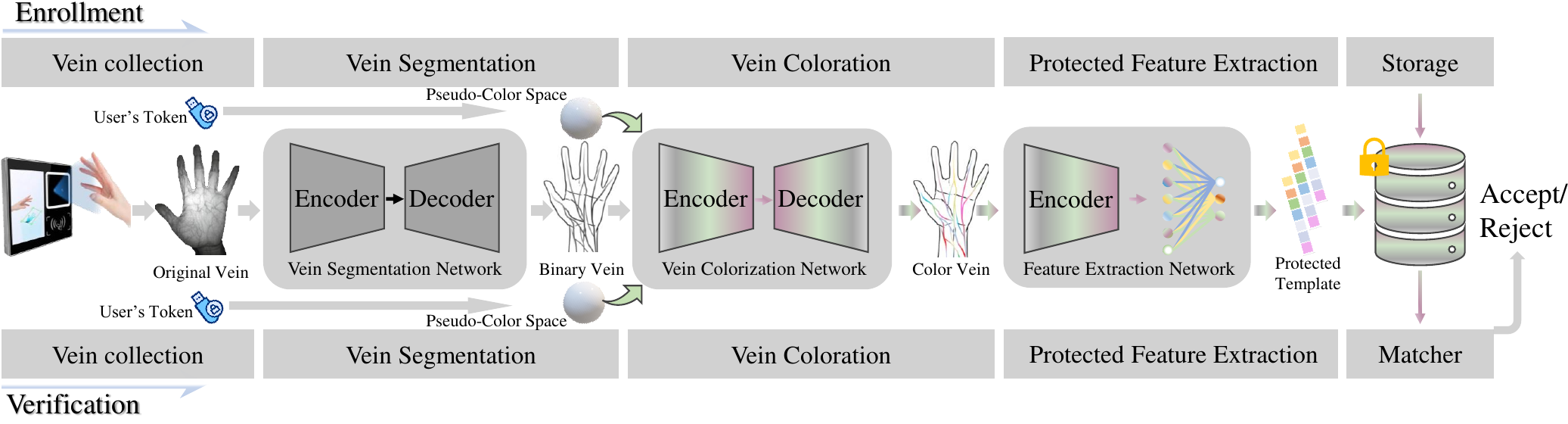}
	\caption{Enrollment/verification procedure of the cancelable biometric system based on the proposed ColorVein scheme.}
	\label{fig2}
\end{figure*}

\subsection{Image Colorization}
Image colorization focus on two main types based on the degree of the user interaction, i.e., guided (or user-guided) and unguided (or data-driven automatic) \cite{vzeger2021grayscale}. Controllability and interactivity are important for image editing, which further enables user defined pseudo-color space using specific tokens for revocability. User-guided methods basically include scribble-based \cite{sangkloy2017scribbler} and example-based \cite{li2017example} methods, while automatic coloring can be equated to deep learning methods. Scribble-based methods guide the colorization process through local color hints provided by the user. Previous work \cite{levin2004colorization, huang2005adaptive} relied heavily on low-level similarity metrics to propagate user-input colors, but typically required extensive user editing to achieve satisfactory results. In recent years, machine learning techniques, especially neural networks \cite{lee2020reference}, have been applied to automatically learn the similarity between pixels, further improving the colorization effect. With the development of deep learning techniques, automated colorization methods \cite{kang2023ddcolor, li2023image, he2023lkat} have emerged. These methods directly learn the mapping from grayscale images to color images by training convolutional neural networks on large-scale image sets. Although these methods can produce realistic results, they usually can only generate a single coloring scheme, which cannot satisfy users' needs for diverse results. This is also important for the diversity (revocability) of cancelable templates.

To combine automatic colorization and user guidance to improve colorization diversity, many studies \cite{deshpande2015learning, deshpande2017learning} have recently began exploring the integration of global and local user inputs into deep learning frameworks to achieve more intuitive and expressive interactive colorization systems. These methods provide a diverse coloring result and also give the user more control over the final visual effect.

\section{Methodology}
\label{sec:ColorVein}
In this section, we introduce the proposed ColorVein scheme. We first introduce the scheme's overall procedure. Then we sequentially introduce vein segmentation, vein colorization, and the generation and matching of ColorVein-based protection templates.

\subsection{Overall Procedure}
ColorVein is a cancelable vein biometrics generation scheme. Fig. \ref{fig2} shows the overall process of using ColorVein to generate cancelable biometrics for enrollment/verification. Generating cancelable biometrics using ColorVein includes three main steps. For user $X$, the $\{I_X,B_X\}$ information pair represents a cancelable identity of $X$. $I_X$ represents the defined identity, and $B_X$ represents $X$'s vein biometric. Firstly, we extract a binarized vein pattern image from the original vein image using vein segmentation network. Next $I_X$ is used to generate a pseudo-random hint $hint(I_X)$ that uniquely corresponds to this defined identity, $hint(I_X)$ and the binarized vein image are fed into a vein colorization network to generate the protected colored vein image. Finally the colored vein image is fed into the feature extraction network to generate vein feature vectors for matching.
\subsection{Vein Segmentation}
\begin{figure}[t]
	\includegraphics[scale=1,width=0.48\textwidth]{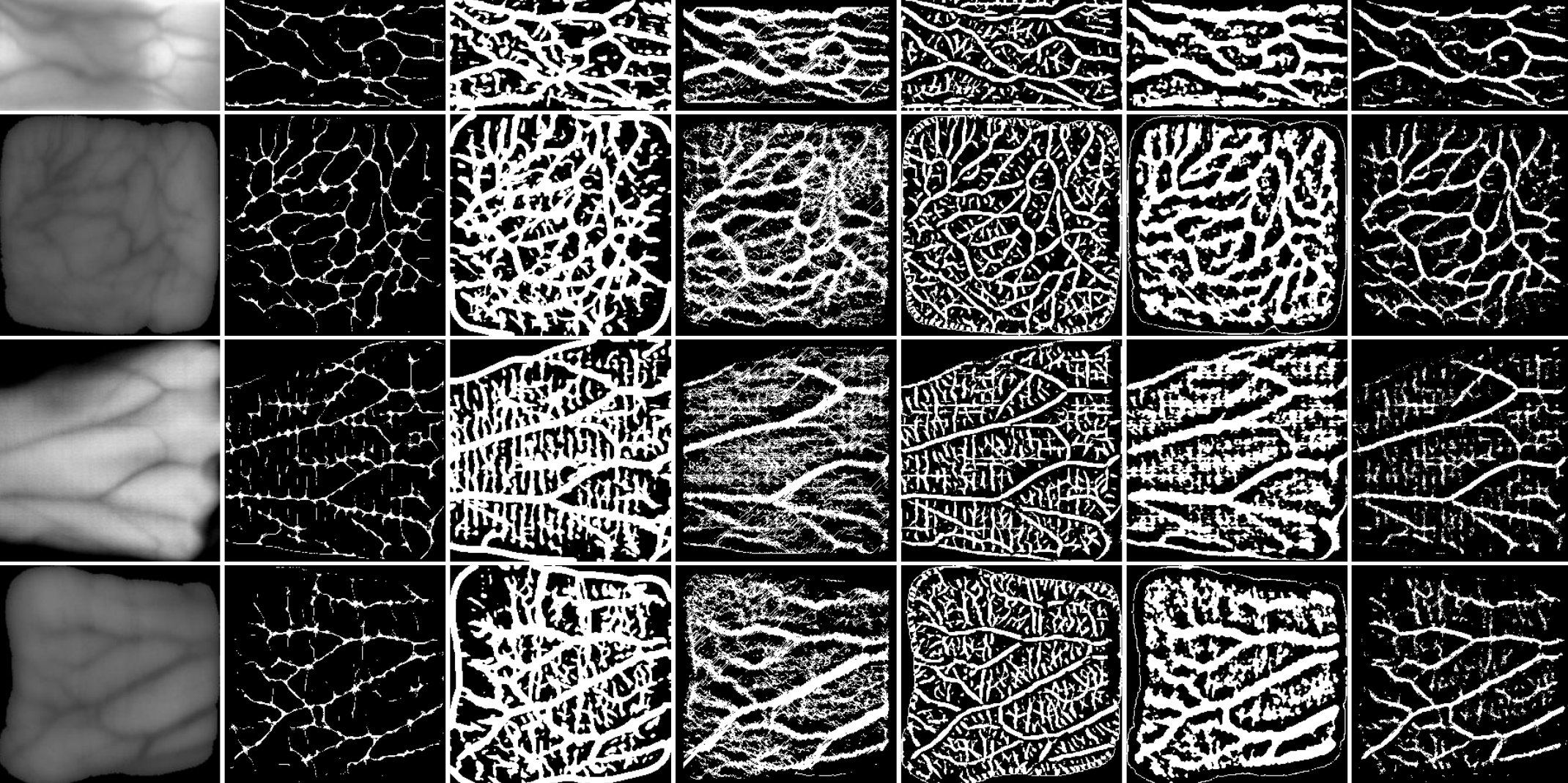}
	\caption{Typical vein pattern extraction methods. Row 1 to 4 are finger, palm, dorsal hand and wrist veins. Column 1 to 7 are the original, MC, PC, RLT, GF, IUTW and fusion image, respectively.}
	\label{fig3}
\end{figure}
Traditional vein pattern feature extraction methods include the maximum curvature (MC) \cite{miura2007extraction}, principal curvature (PC) \cite{choi2009finger}, repeated line tracking (RLT) \cite{miura2004feature}, Gabor filter (GF) \cite{kumar2011human}, and isotropic undecimated wavelet transform (IUWT) \cite{guo1996convolution} etc. Fig. \ref{fig3} shows the binary vein pattern images extracted by several typical methods. Unlike these traditional methods, we use a deep learning model to extract binary vein patterns that is more robust to illumination variations, contrast differences, and noise disturbances, enabling automatic vein pixel labeling. Specifically, we use an encoder-decoder structured deep learning model as the binary vein pattern segmentation tool. The model is able to directly process the original vein images and automatically accomplish the extraction of vein patterns accurately.

We choose ResU-Net \cite{diakogiannis2020resunet} as the vein segmentation network. In order to generate high-quality training labels, we employ a majority voting strategy that fuses the segmentation results of five typical baseline methods (MC, PC, RLT, GF, and IUWT). Specifically, we used the above methods to process vein images separately, where a pixel is classified as a vein pixel (value $1$) if it is labeled as a vein by four or more methods, otherwise it is considered as a background pixel (value $0$) \cite{ou2022gan}. The binary pattern maps generated by this method are used as supervised signals for training the segmentation model. In our proposed ColorVein scheme, the vein image is processed using the trained vein segmentation network both in the enrollment and verification stages to obtain the binarized image, which provides the basis for the subsequent colorization process.
\subsection{Vein Colorization}
Vein colorization is crucial to the ColorVein scheme. Color images can carry richer information than grayscale images. Color is also critical in biometrics, such as in face, iris and gait recognition. However, vein recognition is limited by its inherent imaging principles, and conventional vein features can only be captured as infrared grayscale images. In computer vision, convert grayscale images to color has been an active research field, which can effectively assist classification or segmentation tasks. Inspired by image colorization techniques, we design a method to colorize grayscale vein images. Furthermore, since the color space is editable, different cancelable identities can be defined by interactive colorization using different hint point sets.

\begin{table}[t]
	\centering
	\renewcommand{\arraystretch}{1.2}
	\setlength{\tabcolsep}{1pt} 
	\scriptsize
	\caption{Detailed Architecture of Colorization Network}
	\label{tb1}
	\begin{tabular}{c|c|c|c}
		\toprule
		\textbf{Stage} & \textbf{Layer Configuration} & \textbf{Output Size} & \textbf{Params} \\
		\midrule
		\multicolumn{4}{l}{\textbf{Encoder}} \\
		\midrule
		Input & - & 4 $\times$ H $\times$ W & - \\
		\midrule
		\multirow{1}{*}{E1} & 
		$\begin{bmatrix} 
			\text{Conv}(3{\times}3, 64) + \text{ReLU} \\
			\text{Conv}(3{\times}3, 64) + \text{ReLU}
		\end{bmatrix}$ + BN
		& 64 $\times$ H $\times$ W & 39.4K \\
		\midrule
		\multirow{1}{*}{E2} & 
		$\begin{bmatrix}
			\text{Conv}(3{\times}3, 128) + \text{ReLU} \\
			\text{Conv}(3{\times}3, 128) + \text{ReLU}
		\end{bmatrix}$ + BN
		& 128 $\times$ H/2 $\times$ W/2 & 221.7K \\
		\midrule
		\multirow{1}{*}{E3} & 
		$\begin{bmatrix}
			\text{Conv}(3{\times}3, 256) + \text{ReLU} \\
			\text{Conv}(3{\times}3, 256) + \text{ReLU} \\
			\text{Conv}(3{\times}3, 256) + \text{ReLU}
		\end{bmatrix}$ + BN
		& 256 $\times$ H/4 $\times$ W/4 & 1.47M \\
		\midrule
		\multirow{1}{*}{E4} & 
		$\begin{bmatrix}
			\text{Conv}(3{\times}3, 512) + \text{ReLU} \\
			\text{Conv}(3{\times}3, 512) + \text{ReLU} \\
			\text{Conv}(3{\times}3, 512) + \text{ReLU}
		\end{bmatrix}$ + BN
		& 512 $\times$ H/8 $\times$ W/8 & 5.90M \\
		\midrule
		\multicolumn{4}{l}{\textbf{Dilated Convolution Blocks}} \\
		\midrule
		E5-E6 & 
		$\begin{bmatrix}
			\text{Conv}(3{\times}3, 512, d=2) + \text{ReLU} \\
			\text{Conv}(3{\times}3, 512, d=2) + \text{ReLU} \\
			\text{Conv}(3{\times}3, 512, d=2) + \text{ReLU}
		\end{bmatrix}$ + BN $\times$ 2
		& 512 $\times$ H/8 $\times$ W/8 & 14.2M \\
		\midrule
		E7 &
		$\begin{bmatrix}
			\text{Conv}(3{\times}3, 512) + \text{ReLU} \\
			\text{Conv}(3{\times}3, 512) + \text{ReLU} \\
			\text{Conv}(3{\times}3, 512) + \text{ReLU}
		\end{bmatrix}$ + BN
		& 512 $\times$ H/8 $\times$ W/8 & 7.08M \\
		\midrule
		\multicolumn{4}{l}{\textbf{Decoder}} \\
		\midrule
		D8 & 
		$\begin{bmatrix}
			\text{ConvTranspose}(4{\times}4, 256) \\
			\text{Conv}(3{\times}3, 256) + \text{ReLU} \\
			\text{Conv}(3{\times}3, 256) + \text{ReLU}
		\end{bmatrix}$ + BN
		& 256 $\times$ H/4 $\times$ W/4 & 3.28M \\
		\midrule
		D9 & 
		$\begin{bmatrix}
			\text{ConvTranspose}(4{\times}4, 128) \\
			\text{Conv}(3{\times}3, 128) + \text{ReLU}
		\end{bmatrix}$ + BN
		& 128 $\times$ H/2 $\times$ W/2 & 819.8K \\
		\midrule
		D10 & 
		$\begin{bmatrix}
			\text{ConvTranspose}(4{\times}4, 128) \\
			\text{Conv}(3{\times}3, 128) + \text{LeakyReLU}(0.2)
		\end{bmatrix}$
		& 128 $\times$ H $\times$ W & 350.7K \\
		\midrule
		Output & Conv(1$\times$1, 2) + Tanh & 2 $\times$ H $\times$ W & 258 \\
		\bottomrule
		\multicolumn{3}{l}{\textbf{Total Parameters}} & 34.05M \\
	\end{tabular}
	\vspace{-1mm}
	\begin{tablenotes}
	    \scriptsize
		\item[*] BN: BatchNorm2d; d: dilation rate; Conv: Convolution; ConvTranspose: Transposed Convolution
		\item[†] Skip connections: E3→D8, E2→D9, E1→D10
	\end{tablenotes}
\end{table}

\subsubsection{Colorization Network}
We use a user interactive model \cite{10.1145/3072959.3073703} as ColorVein's colorization network, this model propagates the color of user edits by fusing low-level hints and high-level semantic information learned from large-scale data. 
Specifically, the colorization model uses pre-training on large-scale data to learn a priori knowledge of natural color images, along with user control in a traditional edit propagation framework, using DCNN to map grayscale images and sparse user hints directly to output colors.

The training of the colorization network is performed on a large-scale color image dataset (ImageNet \cite{deng2009imagenet}), with the goal of learning colorization capabilities based on grayscale images and hints. The input includes a grayscale image's lightness $L\in R^{H \times W \times 1}$ ($L^*$ channel) and a user hint $H$. The output is $Y \in R^{H \times W \times 2}$, the color estimate ($a^*$ and $b^*$ channels). The model structure is a type of UNet encoder-decoder structure that can be well used for a variety of condition generation tasks. Table \ref{tb1} shows the detailed structure of the colorization network.

Specifically, the main network is modified from the UNet structure. The network consists of 10 convolutional blocks (conv1-10) that form a deeper encoder-decoder structure. In the encoder part (conv1-4), each block contains 2-3 convolutional-ReLU pairs, and the feature map size is progressively halved while the feature dimension is doubled. In the bottleneck section (conv5-7), a dilated convolution is used to retain more detailed information while maintain the receptive field. The decoder part (conv8-10) progressively restores spatial resolution and reduces feature dimensions. BatchNorm layers are added after each convolutional block to improve training stability. A $1\times 1$ convolution and $tanh$ activation function are used at the end of the network to fit the bounded $a^*b^*$ color space. In addition, conv1-8 layers were fine-tuned using pre-training weights. The objective function used for supervised model learning is defined as

\begin{equation}
	\begin{split}
		L_\delta = \frac{1}{HW} \sum_{i=1}^H \sum_{j=1}^W [
		& \frac{1}{2}(X_{i,j} - Y_{i,j})^2 \mathbb{1}_{|X_{i,j} - Y_{i,j}| < \delta} \, + \\
		& \delta(|X_{i,j} - Y_{i,j}| - \frac{1}{2}\delta) \mathbb{1}_{|X_{i,j} - Y_{i,j}| \geq \delta}]
	\end{split},
\end{equation}
where \( \delta \) is a threshold that serves to control the boundary of the error size. When the error is less than \( \delta \), the loss is the mean squared error (MSE); otherwise, the loss is the mean absolute error (MAE), and $\delta$ is set to 1. Here, \(\mathbb{1}_{\text{condition}}\) indicates that it takes the value 1 when the condition is true and 0 when the condition is false. The colorization model also has a local hint branch network. This network reuses features from the main branch. It connects features from multiple layers of the main branch using a hypercolumn approach. It learns two layer classifiers to predict a probability distribution over output colors $\hat{Z}^{H\times W\times Q}$, where $Q$ is the number of quantized color bins. The model is supervised using cross-entropy loss.
\begin{equation}
	\begin{split}
		L_{CE}(Z, \hat{Z}) = -\sum_{i=1}^H \sum_{j=1}^W \sum_{q=1}^Q 
		& Z_{i,j,q} \log(\hat{Z}_{i,j,q}).
	\end{split}
\end{equation}

\subsubsection{Cancelable Color Vein Generation}
Colorization models need to be trained on a large-scale color image dataset (ImageNet) to learn the a priori knowledge of natural color images, so that the model can diffuse user-edited colors by hint points. Specifically, the generated color veins are obtained by feeding grayscale vein images and a user-defined pseudo-random color space into the trained model. Suppose $G$ is a model trained on the ImageNet dataset, and we inference on $G$ to colorize the vein image. The input includes $L \in R^{H \times W \times 1}$ that is the lightness channel of the input image in LAB color space, and $hint(I_X)$, the pseudo-color space that uniquely corresponds to the defined cancelable $I_X$. The output is a vein color distribution estimate $Y \in R^{H \times W \times 2}$. The colored veins are obtained by combining lightness $L$ with $Y$. In addition to using the set of hint points when defining the pseudo-random color space, it is also possible to set a pseudo-random lightness and vein undertone for the vein region that uniquely corresponds to $I_X$. This is achieved by relying on a special grayscale image type of the input coloring model-binary image (which can be considered as a mask for the vein pattern region).

Feature alignment is important for all cancelable biometric features. To compensate for vein misalignment (translation or rotation, etc.). To ensure stability and accuracy of colorization, we perform ±30 pixel searches horizontally and vertically to maximize the number of hint points within the vein region.

When a template is compromised or stolen, it is easy to define a completely new identity for the user, generating new unique corresponding hint points, lightness, and vein undertones. The revocability of ColorVein will be experimentally verified in Section \ref{sec:revocability}. In addition we evaluate the token stolen scenario, which is verified and discussed through experiments in Section \ref{sec:5B}.
\subsection{Protected Template Generation}
\begin{figure}[t]
	\centering
	\includegraphics[scale=1,width=0.49\textwidth]{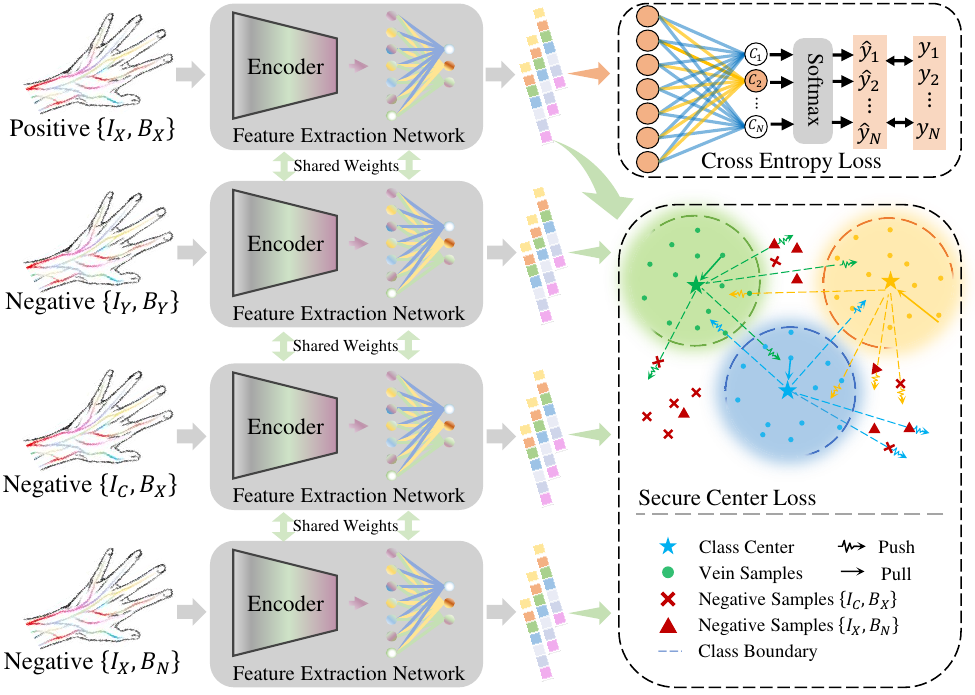}
	\caption{Feature extraction network. $\{I,B\}$ represent an input cancelable vein feature information pairs. $I_X$ and $B_X$ represent enrolled cancelable identities and biometric. $I_Y$ and $B_Y$ represent illegal identities and biometric of impostor. $I_C$ represents cross-application cancelable identities. $B_N$ represents the vein biometrics of the adversary in the stolen scenario.}
	\label{fig4}
\end{figure}
After colorization, colored veins are further feature extracted to obtain a fixed-length feature vector for enrollment/verification. We use ResNet50 \cite{he2016deep} as the backbone of the deep feature extraction network. A fully connected layer is added after layer4 for extracting the depth feature vector with a fixed length of 64. Fig. \ref{fig4} shows the feature extraction network. Unlike conventional biometrics, besides the inherent variability of biometrics, cancelable biometrics needs to additionally consider the revocability, irreversibility and unlinkability of templates. Therefore, instead of using regular recognition/verification loss for biometrics~\cite{hou2021arcvein}, we want to learn a secure extraction function $F$, so we propose the secure center (SC) loss function to supervise the learning of the feature extraction network. Specifically, SC loss is a quintuple loss, the quintuple consists of class centers, enrolled user sample (positive), impostor sample (negative), cross-enrollment sample (negative), and sample generated using a stolen token (negative). Four comparison pairs, including one positive and three negative comparison pairs. The positive comparison pairs are the class center and the positive samples, and the negative comparison pairs are the class center and the other three negative samples, respectively. A great cancelable biometric system should have clearly distinguishable score distributions for genuine matches and impostors. In order to satisfy revocability, the cancelable biometric system needs that the distribution of scores generated by the user's cross-application matches should be the same as that of the impostor. To avoid security risks to the recognition system when the token is stolen (the adversary can access the system using the token with any biometric reference). The cancelable biometric identification system also requires that the match score distribution of the samples generated using the stolen token be the same as the impostor. The SC loss aims to maximize the distances between three types of negative samples and the class center. This ensures consistent matching score distributions for all three types of negative samples. Additionally, it creates a clear separation between these distributions and the genuine matching scores. The loss function used for supervised feature extraction networks in addition includes a $Softmax$ loss. so the total loss is defined as
\begin{equation}
	\small
	\begin{split} 
		L &= L_S + L_{SC} \\
	 &=-\sum_{i=1}^N \log \frac{e^{W_{y_i}^T x_i + b_{y_i}}}{\sum_{k=1}^K e^{W_k^T x_i + b_j}}
		+\sum_{i=1}^N \sum_{k \neq y_i}^K \sum_{j=1}^3 \lambda_j \lfloor D_{i,k}^j + m \rfloor_+,
	\end{split}
\end{equation}
where $L_S$ is the $Softmax$ loss and $L_{SC}$ is the secure center loss. $N$ represents the number of samples in a minibatch. 
$K$ represents the number of classes. $D_{i,k}^j$ represents the difference between the distances of the $i$-th sample and a negative sample to the center of class $y_i$. It is calculated as $D_{i,k}^j = d_{i,y_i} - d_{i,k}^j$. Here, $d_{i,y_i}$ denotes the intra-class distance. The term $d_{i,k}^j$, where $j$ ranges from 1 to 3, represents the distances between $y_i$ and three types of negative samples $\{I_Y,B_Y\}$, $\{I_C,B_X\}$, and $\{I_X,B_N\}$ respectively. M denotes the margin difference. The corresponding $\lambda$ values are set to 1, 0.001, and 0.001. All ColorVein biometric features used for enrollment/verification need to be extracted as fixed-length feature vector templates using a trained feature extraction network, and used for storage or matching.

\section{Experiments and Discussions}
\label{sec:expt}
In this section, we performed experiments on four vein biometrics to evaluate the effectiveness and advancement of ColorVein. This section first introduces the experimental setup, including the datasets, evaluation metrics, and implementation details. Secondly, the recognition performance of a cancelable recognition system designed based on the ColorVein scheme is evaluated, and the effects of different hint point numbers is discussed. Finally ColorVein is compared with state of the art cancelable schemes.
\begin{table}[h]
	\caption{Details of the experimental datasets and training/test set partitioning}
	\renewcommand\arraystretch{1}
	\centering
	\setlength\tabcolsep{1pt}
	\label{tb2}
	\begin{tabular}{ccccc||cccc}
		\toprule
		\multirow{2}{*}{Datasets} & \multirow{2}{*}{Subjects} & \multirow{2}{*}{\begin{tabular}[c]{@{}c@{}}Fingers\\ /Hands\end{tabular}} & \multirow{2}{*}{\begin{tabular}[c]{@{}c@{}}Capture\\ times\end{tabular}} & \multirow{2}{*}{\begin{tabular}[c]{@{}c@{}}Total\\ images\end{tabular}} & \multicolumn{2}{c}{\begin{tabular}[c]{@{}c@{}}Enrollment\\ (Training/Test)\end{tabular}} & \multicolumn{2}{c}{\begin{tabular}[c]{@{}c@{}}Stolen Scenario\\ (Training/Test)\end{tabular}} \\ \cmidrule{6-9} 
		&                           &                                                                           &                                                                          &                                                                         & \multicolumn{1}{l|}{Subjects}                & \multicolumn{1}{l|}{Total}                & \multicolumn{1}{c|}{Subjects}                                  & Total                                 \\ \midrule
		HKPU-FV                   & 156                       & 2                                                                         & 12/6                                                                     & 3132                                                                    & 210                                          & 1260                                      & 51                                                             & 306                                   \\ \midrule
		PUJ-DHV                   & 138                       & 2                                                                         & 4                                                                        & 1104                                                                    & 220                                          & 440                                       & 28                                                             & 112                                   \\ \midrule
		PUT-PV                    & 50                        & 2                                                                         & 12                                                                       & 1200                                                                    & 80                                           & 480                                       & 10                                                             & 120                                   \\ \midrule
		PUT-WV                    & 50                        & 2                                                                         & 12                                                                       & 1200                                                                    & 80                                           & 480                                       & 10                                                             & 120                                   \\ \bottomrule
	\end{tabular}
\end{table}
\subsection{Experimental Setup}
\subsubsection{Datasets}
Four publicly available vein image datasets are used for the experiments, including a finger vein dataset HKPU-FV released by the Hong Kong Polytechnic University \cite{kumar2011human}, a palm vein dataset PUT-FV and a wrist vein dataset PUT-WV released by Poznań Polytechnic University \cite{kabacinski2011vein}, and a dorsal hand vein dataset PUJ-DHV released by Benha University \cite{9606904}. HKPU-FV extracts ROIs using the method proposed by \cite{yao2020robust} and the size is normalized to $128\times 256$. PUT-FV, PUT-WV and PUJ-DHV extract ROIs using the maximum outside bounding rectangle and the size is normalized to $256\times 256$. Table \ref{tb2} shows details of the experimental datasets and training/test set partitioning.

\subsubsection{Evaluation Metrics}
Same as previous work, in this paper the \textbf{\textit{Match Score}} of vein feature vectors is evaluated using cosine similarity,
\begin{equation}
	Match\text{\,} Score = \frac{A \cdot B}{\|A\| \|B\|} = \frac{\sum_{i=1}^{n} A_i B_i}{\sqrt{\sum_{i=1}^{n} A_i^2} \sqrt{\sum_{i=1}^{n} B_i^2}}.
\end{equation}

To evaluate recognition performance, commonly used metrics \textit{equal error rate} (\textbf{\textit{EER}}), which is the point where false accept rate equals to false reject rate.

Irreversibility of cancelable biometrics is quantitatively assessed using the privacy leakage rate \cite{pinto2020secure}:
\begin{equation}
	\frac{H(X | Y)}{H(X)} = 1 - \frac{I(X; Y)}{H(X)},
\end{equation}
where $X$ is the original biometric, $Y$ is the cancelable template, $H(X)$ denotes the entropy of $X$, $H(X|Y)$ represents the conditional entropy of $X$ given $Y$, and $I(X; Y)$ denotes the mutual information between $X$ and $Y$. The privacy leakage rate should be as high as possible: even if the adversary is fully known about $Y$, getting information about $X$ should be impossible.

Unlinkability of a cancelable biometrics is assessed based on the
following \textit{global unlinkability} $\boldsymbol{D}_\leftrightarrow^{\boldsymbol{sys}}$ metric 
\cite{gomez2017general}:
\begin{equation}\label{eq6}	
	D_\leftrightarrow^{sys}=\int p(s|H_m)[p(H_m|s)-p(H_{nm}|s)]ds,
\end{equation}
where 
$s$ is the linkage score between 
two protected templates $T_1$ and $T_2$.
Based on the paired hypothesis
$H_m=$``both templates belong to mated
instances" and the unpaired hypothesis $H_{nm}=$``both templates belong
to\ non-mated instances", we obtain 
the conditional probabilities
$p(H_m|s)$ and $p(H_{nm}|s)$, 
for a given matching score $s$. $p(s|H_m)$ is the conditional probability of
obtaining a score $s$ knowing that the two templates come from mated instances. A small $D_\leftrightarrow^{sys}$ indicates that the link is weak, and vice versa.

The recognition system is cancelable if the impostor and pseudo-impostor
distributions overlap and the genuine and pseudo-impostor distributions are
clearly distinguishable. The degree of overlap between two distributions
can be measured using the \textit{decidability index} $\boldsymbol{d}^\prime$, which
reflects the revocability of the recognition system
\cite{sadhya2019generation}. This metric is described as the normalized
distance between the means of the two distributions. 
For two distributions with means $\mu_1$ and $\mu_2$, and standard deviations $\sigma_1$ and $\sigma_2$, respectively, $d^\prime$ is defined as
\begin{equation}\label{eq7}	
	d^\prime = \frac{|\mu_1 - \mu_2|}{\sqrt{\frac{1}{2}(\sigma_1^2 +
			\sigma_2^2)}},
\end{equation}
a higher value of $d^\prime$ indicates good separation between the two distributions, and vice versa.
\subsubsection{Implementation Details}
We split the dataset into enrollment part and stolen scenario part, and each part is further divided into training set and test set.
Take the HKPU-FV dataset as an example, it contains 312 subjects. Among them, 210 subjects are designated as legitimate users for enrollment, 6 vein samples of each subject are used for enrollment/training, and another 6 samples are used for verification/testing. The remaining 102 subjects are considered as illegal users for simulating token stolen scenarios, where 51 subjects are used for training and another 51 subjects are used for testing. Table \ref{tb2} details the specific division of the four datasets.

In ColorVein framework, segmentation network and feature extraction network are required for training. We adopt Adam \cite{kingma2014adam} as the optimization method, with momentum set to 0.9, learning rate set to 0.001. They are trained on an RTX 4090 GPU, Intel(R) Xeon(R) Gold 5218R CPU @ 2.10GHz processor, and 440GB RAM, and is implemented using the PyTorch framework. Note that the Segmentation network is trained 200 epochs with batch size of 32, while the feature extraction network is trained 500 epochs with batch size of 8.

\subsection{Effect of Hint Point Number $m$}
\label{sec:5B}
\begin{figure}[]
	\includegraphics[scale=1,width=0.48\textwidth]{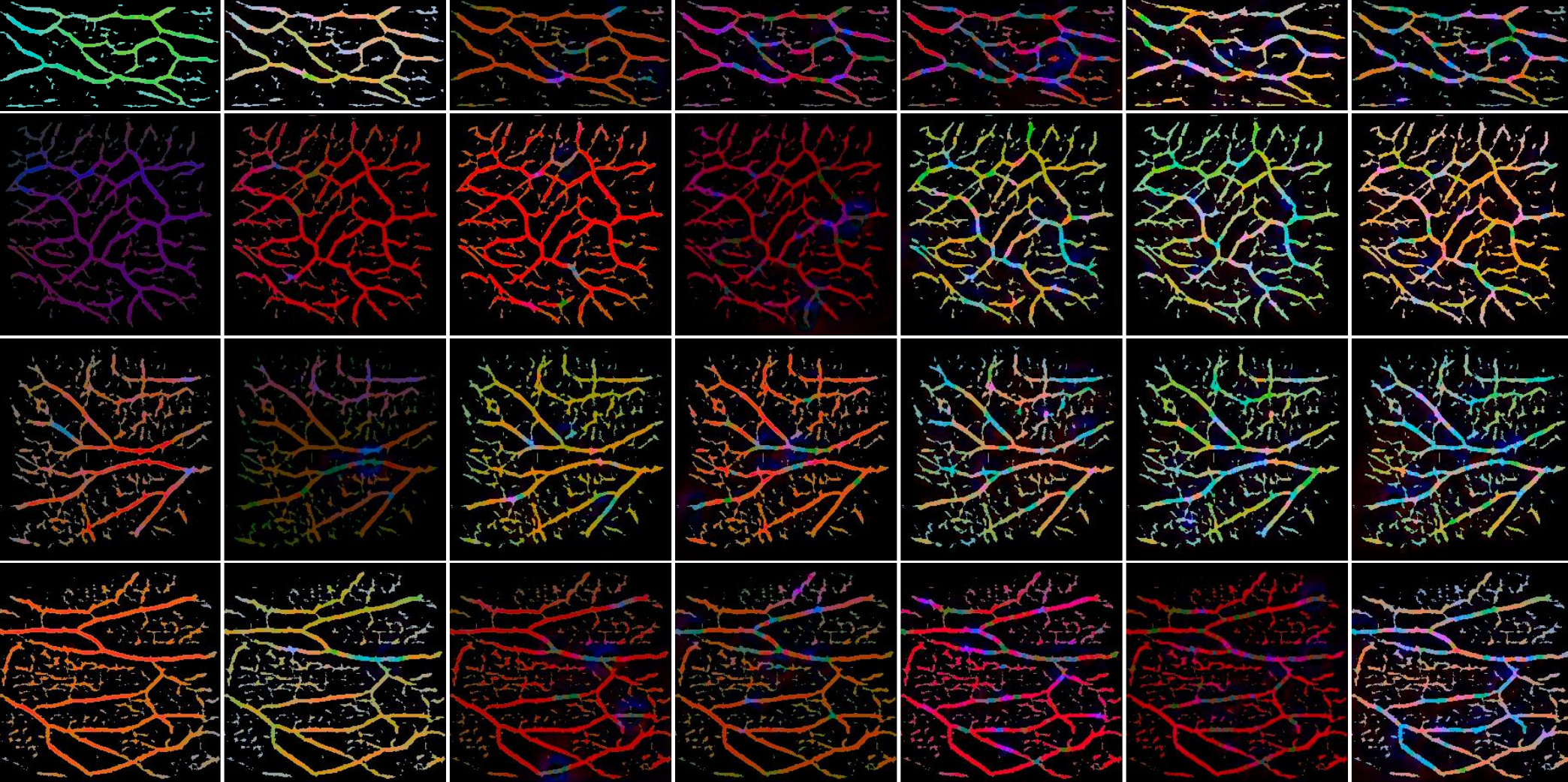}
	\caption{Vein image after colorization by the colorization network. Row 1 to 4 are finger, palm, dorsal hand and wrist veins. Columns 1 to 7 indicate the ColorVein hint point numbers of 1, 5, 10, 25, 50, 75 and 100, respectively.}
	\label{fig5}
\end{figure}

Hint point numbers $m$ will affect the colorization results of the ColorVein scheme, so we evaluated the effects of different $m$ on ColorVein performance. From a visual perspective, the most direct effect of the number of hint points is reflected in the color richness of the ColorVein template. As the hint point numbers increases, the vein region is marked by more different colors, spreading over the region around the vein through the colorization network. Fig. \ref{fig5} shows ColorVein templates generated using different numbers of hint points. In addition, injecting richer color information into the veins effectively increases the interclass distance of the vein samples.

\begin{figure}[t]
	\centering
	\includegraphics[scale=1,width=0.42\textwidth]{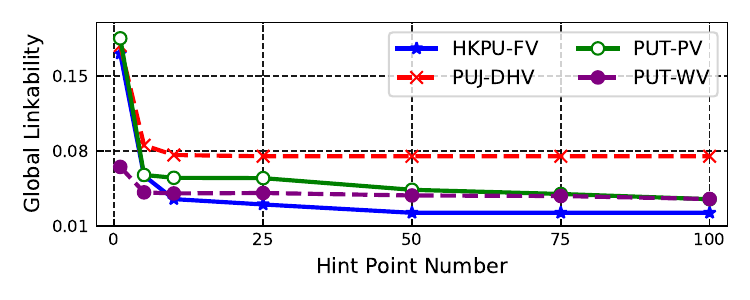}
	\caption{The effect of different hint point numbers on global linkability.}
	\label{fig6}
\end{figure}

\begin{table}[]
	\centering
	\renewcommand\arraystretch{0.5}
	\setlength\tabcolsep{1.8pt}
	\caption{EERs (\%) of different hint point number in normal and stolen scenarios. $hint(m)$ represents the hint point number is $m$}
	\label{tb3}
	\begin{tabular}{lcccccccc}
		\toprule
		\multirow{2}{*}{} & \multicolumn{2}{c}{HKPU-FV}                               & \multicolumn{2}{c}{PUJ-DHV}                               & \multicolumn{2}{c}{PUT-PV}                                & \multicolumn{2}{c}{PUT-WV}           \\ \cmidrule{2-9} 
		& \multicolumn{1}{c|}{Normal} & \multicolumn{1}{c|}{Stolen} & \multicolumn{1}{c|}{Normal} & \multicolumn{1}{c|}{Stolen} & \multicolumn{1}{c|}{Normal} & \multicolumn{1}{c|}{Stolen} & \multicolumn{1}{c|}{Normal} & Stolen \\ \midrule
		Baseline          & 2.250                       & -                           & 1.805                       & -                           & 2.699                       & -                           & 8.355                       & -      \\ \midrule
		Binary            & 5.513                       & -                           & 2.186                       & -                           & 3.689                       & -                           & 6.326                       & -      \\ \midrule
		$hint(1)$                 & 0.987                       & 1.460                       & 0.909                       & 0.909                       & 0.832                       & 1.465                       & 1.332                       & 1.751  \\ \midrule
		$hint(5)$                 & 0.875                         & 1.113                         & 0.910                       & 0.910                       & 0.775                       & 1.002                       & 1.208                       & 1.638  \\ \midrule
		$hint(10)$                & 0.797                       & 0.817                       & 0.909                       & 0.909                       & 0.667                       & 0.770                       & 1.085                       & 1.000  \\ \midrule
		$hint(25)$                & 0.792                       & 0.792                       & 0.909                       & 0.909                       & 0.664                       & 0.664                       & 0.670                       & 0.912  \\ \midrule
		$hint(50)$                & 0.635                       & 0.666                       & 0.905                       & 0.905                       & 0.332                       & 0.415                       & 0.584                       & 0.833  \\ \midrule
		$hint(75)$                & 0.635                       & 0.666                       & 0.908                       & 0.908                       & 0.258                       & 0.416                       & 0.583                       & 1.106  \\ \midrule
		$hint(100)$               & 0.476                       & 0.674                       & 0.908                       & 0.908                       & 0.258                       & 0.664                       & 0.507                       & 1.573  \\ \bottomrule
	\end{tabular}
\end{table}

Specifically, we evaluated and analyzed the recognition system based on the original vein image (Baseline), binary vein image, and ColorVein scheme with the number of hint points 1, 5, 10, 25, 50, 75, and 100. We also evaluated the EER of the recognition system under the token stolen scenario. Table \ref{tb3} lists the EERs for each dataset for both normal and stolen scenarios. By analyzing it, we can conclude:

\begin{enumerate}
	\item The original vein image can perform the basic recognition task, but the recognition performance is not satisfactory. When using binary vein images, the performance is significantly reduced, this is mainly due to the fact that the vein segmentation process transforms continuous real-valued samples into binary feature samples, resulting in the loss of information. While ColorVein significantly improves recognition performance by injecting color information into the vein.
	
	\item With the increase of hint point number, recognition system's EER shows a gradual decreasing trend. When the hint point number $m$ is more than 10, the decreasing trend of EER flattens out, which indicates that the carrying capacity of the vein region for color richness has reached the saturation point. Continuing to increase the hint point numbers, the EER shows only a small decrease or remains stable.
	
	\item Using more hint points to define different pseudo-color spaces for the same original vein samples, it increases the difference between them and thus enhances the unlinkability. Fig. \ref{fig6} shows the curves of global linkability with different hint point numbers on each dataset.
	
	\item It is important to note that when the hint point count is more than 10, the EER of the recognition system cannot continue to be significantly reduced, and may have a negative effect. In the stolen scenario, the EER initially decreases as the hint point number increases, but when the hint point number is too high ($m>25$), ColorVein begins to show a dependence on a specific color space. This leads the EER increase in the stolen scenario, so the hint point number should not be too large.
\end{enumerate}

\subsection{Recognition Performance Comparison}
\label{sec:RPC}

We compared ColorVein with several representative cancelable biometrics generation schemes, including block remapping \cite{ratha2001enhancing}, mesh warping \cite{10620353}, Biohashing \cite{jin2004biohashing} and Bloom Filter \cite{bloom}. These schemes have been proven to have feasibility on vein biometrics. Table \ref{tb4} shows the recognition performance of ColorVein compared to these methods in normal and stolen scenarios. The results show that ColorVein has the best recognition performance, with a significant improvement over baseline. Although the other methods have better recognition performance than baseline in a few cases, they cannot effectively ensure the security of the system when the token is stolen. ColorVein is the only scheme that significantly outperforms baseline in recognition performance, and it can clearly distinguish between templates generated using stolen tokens. Analyzing other methods, it is not difficult to find the reason, they destroy the pattern features by block division, distortion, or even project the features into discrete binary space, which leads to the loss of feature information. While ColorVein not only preserves the pattern features, it creatively injects color information into the vein image, enriching its feature information. Therefore, ColorVein is better than other state-of-the-art cancelable biometrics schemes. In addition, we provide a further comparative analysis of ColorVein's unlinkability, revocability, and irreversibility with other schemes in Section \ref{sec:cbac}.
\begin{table}[]
	\setlength\tabcolsep{0.5pt}
	\centering
	\caption{EERs (\%) comparison with representative cancelable biometric schemes in normal and stolen scenarios}
	\label{tb4}
	\begin{tabular}{ccccccccc}
		\toprule
		\multirow{2}{*}{} & \multicolumn{2}{c}{HKPU-FV}                               & \multicolumn{2}{c}{PUJ-DHV}                               & \multicolumn{2}{c}{PUT-PV}                                & \multicolumn{2}{c}{PUT-WV}           \\ \cmidrule{2-9} 
		& \multicolumn{1}{c|}{Normal} & \multicolumn{1}{c|}{Stolen} & \multicolumn{1}{c|}{Normal} & \multicolumn{1}{c|}{Stolen} & \multicolumn{1}{c|}{Normal} & \multicolumn{1}{c|}{Stolen} & \multicolumn{1}{c|}{Normal} & Stolen \\ \midrule
		Baseline          & 2.250                       & -                           & 1.805                       & -                           & 2.699                       & -                           & 8.355                       & -      \\ \midrule
		Binary            & 5.513                       & -                           & 2.186                       & -                           & 3.689                       & -                           & 6.326                       & -      \\ \midrule
		Bloom Filters     & 13.54                       & 0.0                         & 27.58                       & 0.0                         & 13.92                       & 0.0                         & 29.49                       & 0.0    \\ \midrule
		Mesh Warping      & 3.168                       & 0.295                       & 20.61                       & 17.27                       & 16.73                       & 8.563                       & 16.54                       & 14.25  \\ \midrule
		Block Rmapping    & 9.854                       & 6.315                       & 18.04                       & 11.45                       & 4.610                       & 1.938                       & 4.925                       & 1.979  \\ \midrule
		Biohashing        & 6.425                       & 6.315                       & 1.392                       & 25.36                       & 4.921                       & 43.39                       & 2.789                       & 43.39  \\ \midrule
		ColorVein         & 0.797                       & 0.817                       & 0.909                       & 0.909                       & 0.667                       & 0.770                       & 1.085                       & 1.000  \\ \bottomrule
	\end{tabular}
\end{table}

\subsection{Recognition Performance on Lower Quality Samples}
Vein image acquisition is affected by the environment, which may lead to image quality degradation. Therefore, we evaluated the performance of the ColorVein scheme on low-quality vein image samples. In this experiment, we utilize data augmentation methods to generate a test set of low-quality vein images. Specifically, these data augmentation methods include motion blur, skin scattering blur, optical blur, over/under-exposure~\cite{choi2021restoration, choi2020modified}. Finally, we produced 10,000 low-quality image samples on four datasets and used these samples to test the robustness of the ColorVein scheme to low-quality image samples. Table \ref{tb5} lists the EERs obtained by the ColorVein scheme in low-quality image samples on each dataset.
\begin{table}[]
	\renewcommand\arraystretch{0.5}
	\centering
	\caption{EERs(\%) of low quality samples on each dataset}
	\label{tb5}
	\begin{tabular}{cccc}
		\toprule
		HKPU-FV & PUJ-DHV & PUTP-PV & PUT-WV \\ \midrule
		1.701   & 1.958   & 0.506   & 0.562  \\ \bottomrule
	\end{tabular}
\end{table}

\begin{figure}[t]
	\centering
	\includegraphics[scale=1,width=0.3\textwidth]{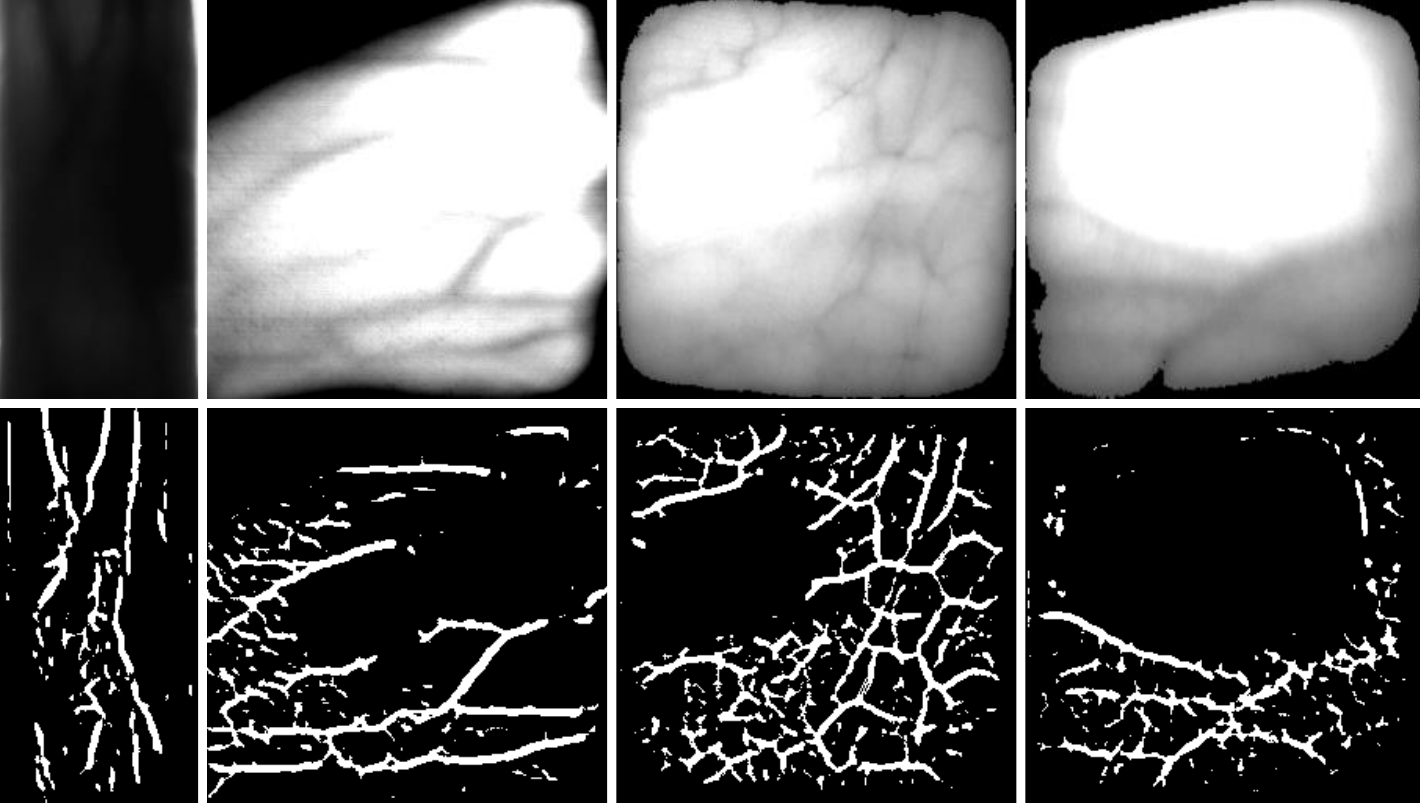}
	\caption{Low quality vein image samples and its binary pattern features.}
	\label{fig7}
\end{figure}

Observing Table \ref{tb5}, we find that the performance of the ColorVein scheme remains stable or slightly increases on low-quality images, which proves that ColorVein has good robustness to low-quality samples. We further observe the mis-matched samples, which generally exhibit extremely low quality. As shown in Fig. \ref{fig7}, when the image is over/under-exposed, the vein pattern cannot be imaged effectively, resulting in vein pattern information being completely lost. In such samples, a valid vein pattern cannot be extracted, this phenomenon is detrimental to all vein recognition tasks and can lead to obvious false matches. In vein recognition/verification tasks, this phenomenon can be mitigated by vein pattern reconstruction or recovery~\cite{qin2017deep}.

\begin{table}[h]
	\centering
	\renewcommand\arraystretch{0.5}
	\setlength\tabcolsep{2.2pt}
	\scriptsize
	\caption{Time cost of template generation for each cancelable scheme}
	\label{tb6}
	\begin{tabular}{cccccc}
		\toprule
		\multicolumn{1}{c|}{Method} & \multicolumn{1}{c|}{Block Remaping} & \multicolumn{1}{c|}{Mesh Warpping} & \multicolumn{1}{c|}{Bloom Filter} & \multicolumn{1}{c|}{Biohashing} & ColorVein \\ \midrule
		Time/ms                     & 8.140                                & 10.12                              & 78.01                             & 6.305                           & 21.77     \\ \bottomrule
	\end{tabular}
\end{table}
\subsection{Cancelable Template Generation Time Analysis}
We compared the time required to generate cancelable templates for the ColorVein scheme with several representative schemes. Table \ref{tb6} presents the average template generation time for each method across four datasets. It is important to note that, due to the different input formats required by each method, the listed time includes the process of extracting original biometric features. As shown in Table \ref{tb6}, the generation time of cancelable templates for the ColorVein scheme is only 21.77 ms, which is slightly higher than that of the Block Remapping, Mesh Warping, and Biohashing schemes, and lower than the Bloom Filter. This result indicates that ColorVein has low deployment complexity and real-time processing potential while ensuring biometric template protection.

\begin{figure}[t]
	\centering
	\includegraphics[scale=1,width=0.28\textwidth]{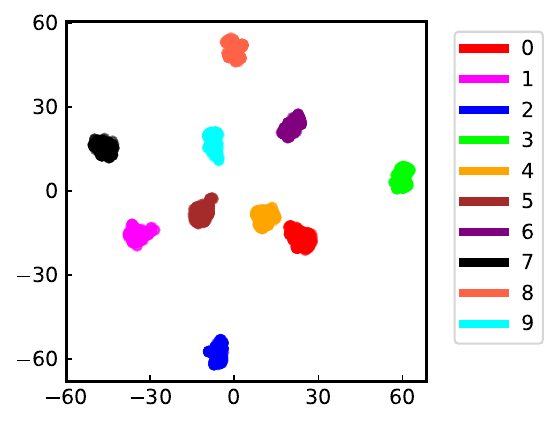}
	\caption{Visualization of the 2D Feature Distribution of ColorVein.}
	\label{fig8}
\end{figure}

\subsection{Feature Visualization}
We visualized the feature distributions of the ColorVein scheme in the HKPU-FV dataset. Specifically, we randomly selected 10 classes and augmented the number of test samples for each class to 600 by traditional data augmentation methods. Subsequently, we analyzed the feature distributions of these test samples, and Fig. \ref{fig8} shows the results of the visualization of the 2D feature distributions. The results show that the class distinctions between the samples are very clear, while the samples of the same class tend to be concentrated in the center of their respective classes, with minimal inter-class gaps.

\section{Security and Privacy Analysis}
\label{s&p}
In this section, we further analyzed the ColorVein-based cancelable vein recognition system, including privacy, security, unlinkability, and revocability analysis. We also qualitatively compare ColorVein with existing representative cancelable biometric schemes, including recognition performance, irreversibility, unlinkability, and revocability.
\begin{table}[h]
	\renewcommand\arraystretch{0.5}
	\centering
	\setlength\tabcolsep{2pt}
	\caption{Privacy leakage rate on each dataset}
	\label{tb7}
	\begin{tabular}{lcccc}
		\toprule
		& HKPU-FV & PUJ-DHV & PUT-PV & PUT-WV \\ \midrule
		Garyscale Vein & 0.951   & 0.960   & 0.962  & 0.954  \\ \midrule
		Binary Vein    & 0.987   & 0.985   & 0.991  & 0.988  \\ \bottomrule
	\end{tabular}
\end{table}
\subsection{Privacy Analysis}
\subsubsection{Irreversibility}

Irreversibility refers to that it is not feasible to reconstruct the original biometric feature from the protected feature template. We evaluated the privacy-preserving effect of ColorVein cancelable biometric templates on original grayscale vein and binary vein images by privacy leakage rate. Table \ref{tb7} shows the privacy leakage rate on each dataset. The results show that the privacy leakage rate is generally high and close to 1, it means that even if the adversary has all the information about ColorVein templates, they cannot obtain valid information about the original biometric, thus unable to reconstruct the original biometrics from the protected templates. It is worth noting that recent studies have shown that optimized deep learning models are able to automatically compress the inputs, retaining only the critical information needed to accomplish a specific task \cite{pinto2020secure}. This means that we use DCNN-based feature extraction network, which may achieve good irreversibility even without relying on well-designed feature extraction algorithms.

\subsubsection{Attacks via Record Multiplicity for Privacy}
We also consider an RM attack on recovering the original biometric feature. The RM attack attempts to reconstruct the original biometric feature template using multiple lost template instances, those templates may or may not contain knowledge and parameters relevant to the reconstruction algorithm. ColorVein transforms the template to a rank space \cite{2017ranking} that is not relevant to the original feature space, and even if the adversary obtains multiple compromised ColorVein templates, it is unable to effectively reconstruct the original biometric feature.
\begin{figure}[t]
	\captionsetup[subfloat]{labelsep=none,format=plain,labelformat=empty}
	\subfloat[\hspace{1.4em} \scriptsize{(a) HKPU-FV}]{
		\includegraphics[width=0.45\linewidth]{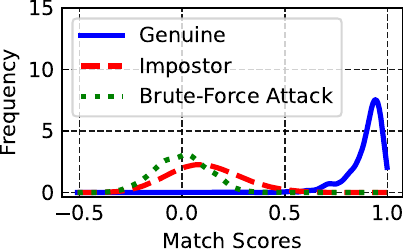}}
	\subfloat[\hspace{1.8em}\scriptsize{(b) PUT-PV}]{
		\includegraphics[width=0.45\linewidth]{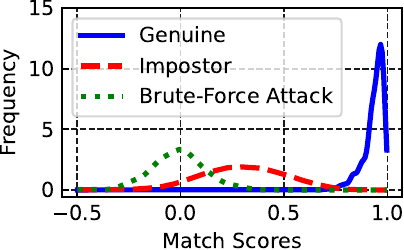}}
	\vspace{-10pt}
	\subfloat[\hspace{1.6em}\scriptsize{(c) PUJ-DHV}]{
		\includegraphics[width=0.45\linewidth]{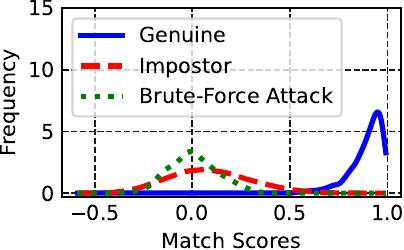}}
	\subfloat[\hspace{1.8em}\scriptsize{(d) PUT-WV}]{
		\includegraphics[width=0.45\linewidth]{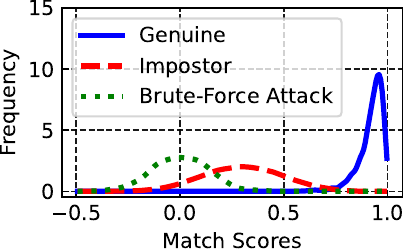}}
	\caption{The Genuine, Impostor and Brute Force Attack match score distribution on each dataset.}
	\label{fig9} 
\end{figure}
\subsection{Security Analysis}
\subsubsection{Brute Force Attack}
Brute force attack, it refers to an adversary generates legitimate biometric templates to access the recognition system by exhaustively enumerating or speculating possible biometric data. The ColorVein generated features range from $[-10,10]$ and have a fixed precision of four decimals, that means $200,000$ $(\approx2^{17})$ attempts are needed for each feature component. Thus, fully inferring the 64-dimensional feature vector requires $2^{(17 \times 64)}=2^{1088}$ attempts, which is computationally infeasible. To verify this, we randomly generated 10,000 templates to attack the identification system and plotted the distribution of brute force attack match scores. Fig. \ref{fig9} shows match score distributions of brute force attacks and genuine/impostor match score distributions in normal scenarios. The results show that the distribution of match scores generated by brute force attacks either overlaps with the impostor distribution or shows a leftward trend, showing lower match scores. The EER on all datasets is 0, indicating that the distribution of brute force attack matching scores is completely different from genuine matching scores, which can be completely distinguished by the system. It is proved that ColorVein can effectively defend against brute force attacks.

\subsubsection{False Accept Attack}
\begin{figure}[t]
	\centering
	\captionsetup[subfloat]{labelsep=none,format=plain,labelformat=empty,skip=-10pt}
	\subfloat[\hspace{1.4em} \scriptsize{HKPU-FV}]{
		\includegraphics[width=0.7\linewidth]{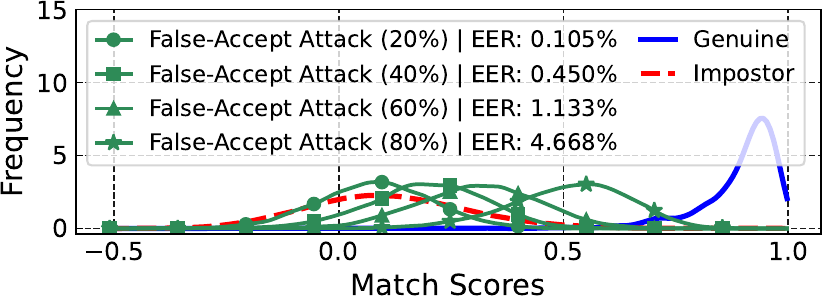}}
	
	\vspace{-10pt}
	
	\subfloat[\hspace{1.8em}\scriptsize{PUT-PV}]{
		\includegraphics[width=0.7\linewidth]{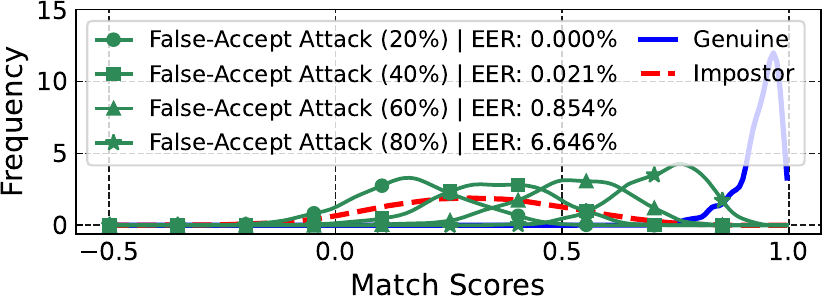}}
	
	\vspace{-10pt}
	
	\subfloat[\hspace{1.6em}\scriptsize{PUJ-DHV}]{
		\includegraphics[width=0.7\linewidth]{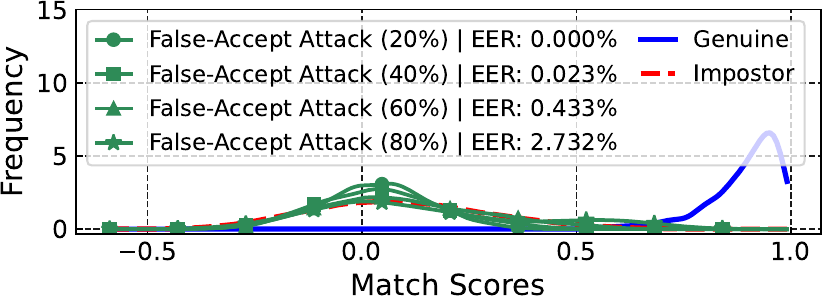}}
	
	\vspace{-10pt}
	
	\subfloat[\hspace{1.8em}\scriptsize{PUT-WV}]{
		\includegraphics[width=0.7\linewidth]{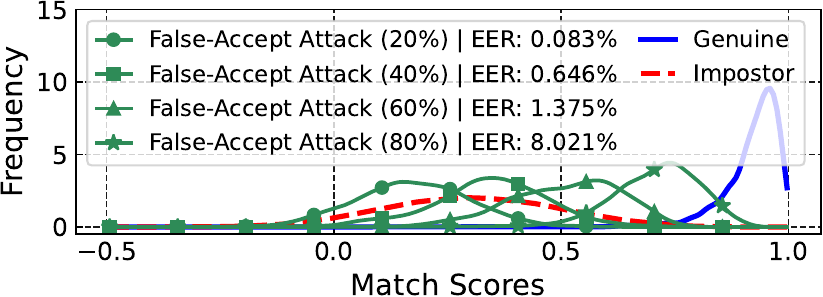}}
	
	\caption{The Genuine, Impostor and False Accept Attack match score distribution on each dataset. False Accept Attack ($N\%$)| EER: $n\%$, where $N\%$ indicates the proportion of correctly guessed bits, and $n\%$ indicates the EER between the False Accept Attack score and the genuine score at $N\%$ proportion.}
	\label{fig10} 
\end{figure}

False accept attack differ from brute force attack as it might need less attempts to get access. This attack assumes the adversary fully knows the template protection process of the identification system, thus increases the possibility of forging legitimate templates. In the threshold-based decision scheme, the match score of the forged template just needs to exceed a predetermined threshold to successfully access the system. To evaluate this attack, we assume that the adversary successfully speculates the $N$\% bits in ColorVein and tries to access the identification system. Fig. \ref{fig10} shows the distribution of false accept attack and impostor match scores. It can be observed that as more bits are speculated, the false attack matching scores gradually move closer to the genuine matching score distribution. The results show that only when $N$ is greater than $60$\% the adversary has a very small chance to get access. This means that at least 38 bits of feature vectors need to be successfully inferred, which is equivalent to $2^{(17 \times 38)} = 2^{646}$ attempts. Thus false accept attack is still computationally infeasible.

\subsubsection{Attacks via Record Multiplicity for Security}
RM security attacks falsify legitimate templates by collecting multiple protected templates of the same biometric identity enrolled in different applications, so they are also called correlation or federation attacks. As the name suggests, the adversary first needs to obtain linked templates, which means linking different protected templates generated from the same finger vein template. We evaluated and analyzed unlinkability in Section \ref{sec:unlinkability}. Based on the fact that ColorVein exhibits very low linkability, it can effectively defend such attacks.
\subsection{Unlinkability}
\label{sec:unlinkability}
\begin{figure}[t]
	\captionsetup[subfloat]{labelsep=none,format=plain,labelformat=empty}
	\subfloat[\hspace{0.8em} \scriptsize{(a) HKPU-FV}]{
		\includegraphics[width=0.45\linewidth]{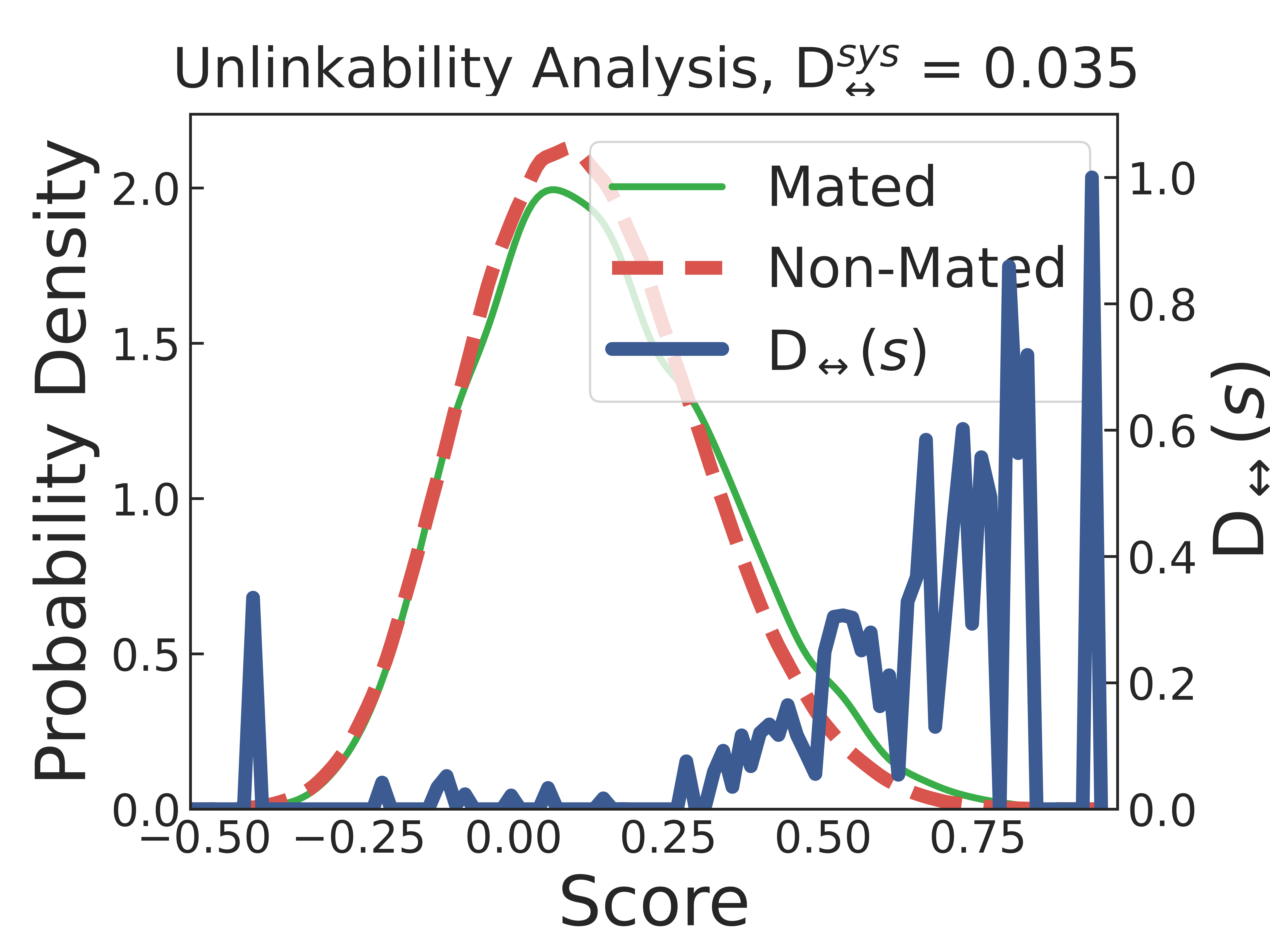}}
	\subfloat[\hspace{0.8em}\scriptsize{(b) PUT-PV}]{
		\includegraphics[width=0.45\linewidth]{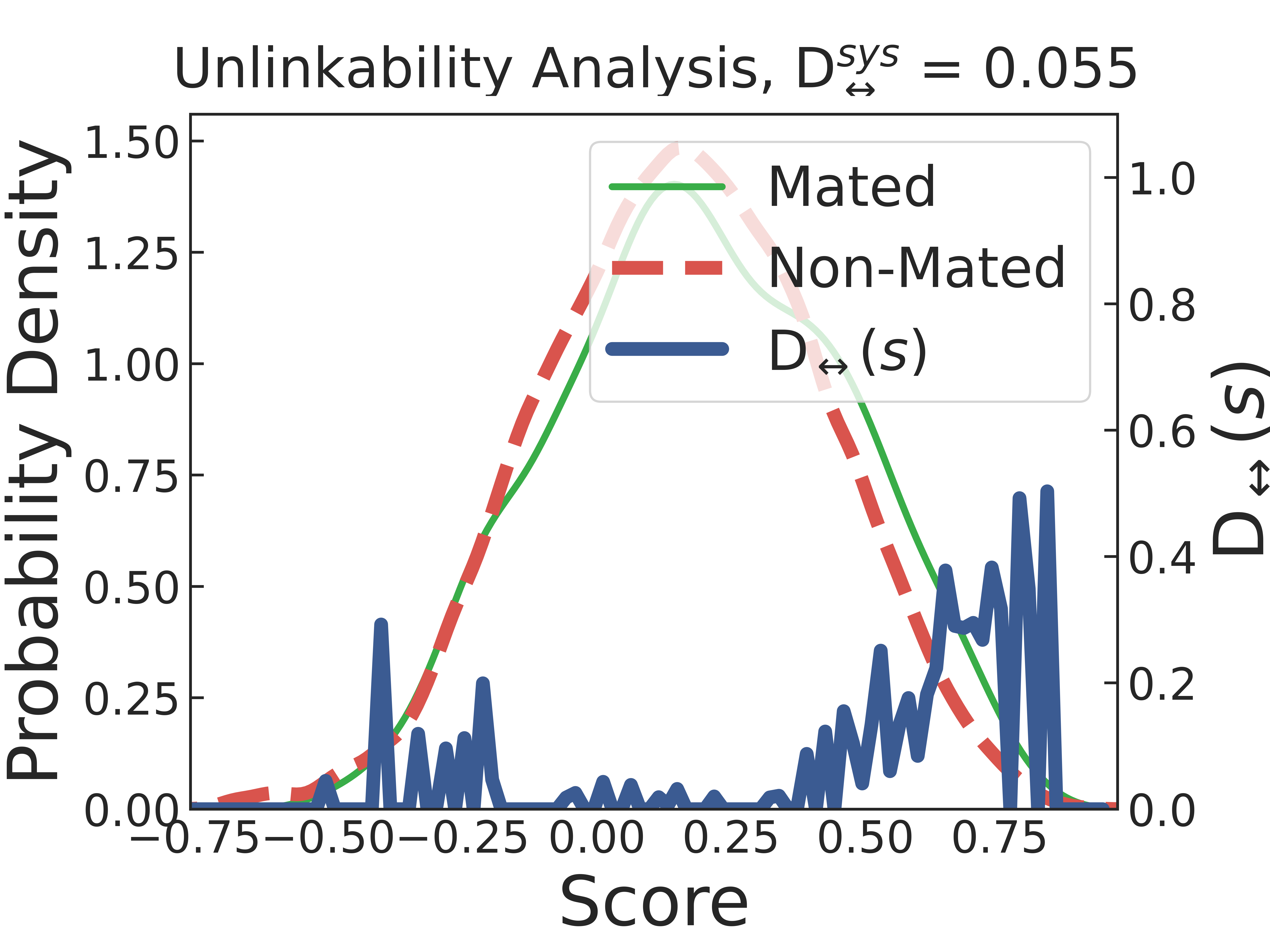}}
	\vspace{-10pt}
	\subfloat[\hspace{0.8em}\scriptsize{(c) PUJ-DHV}]{
		\includegraphics[width=0.45\linewidth]{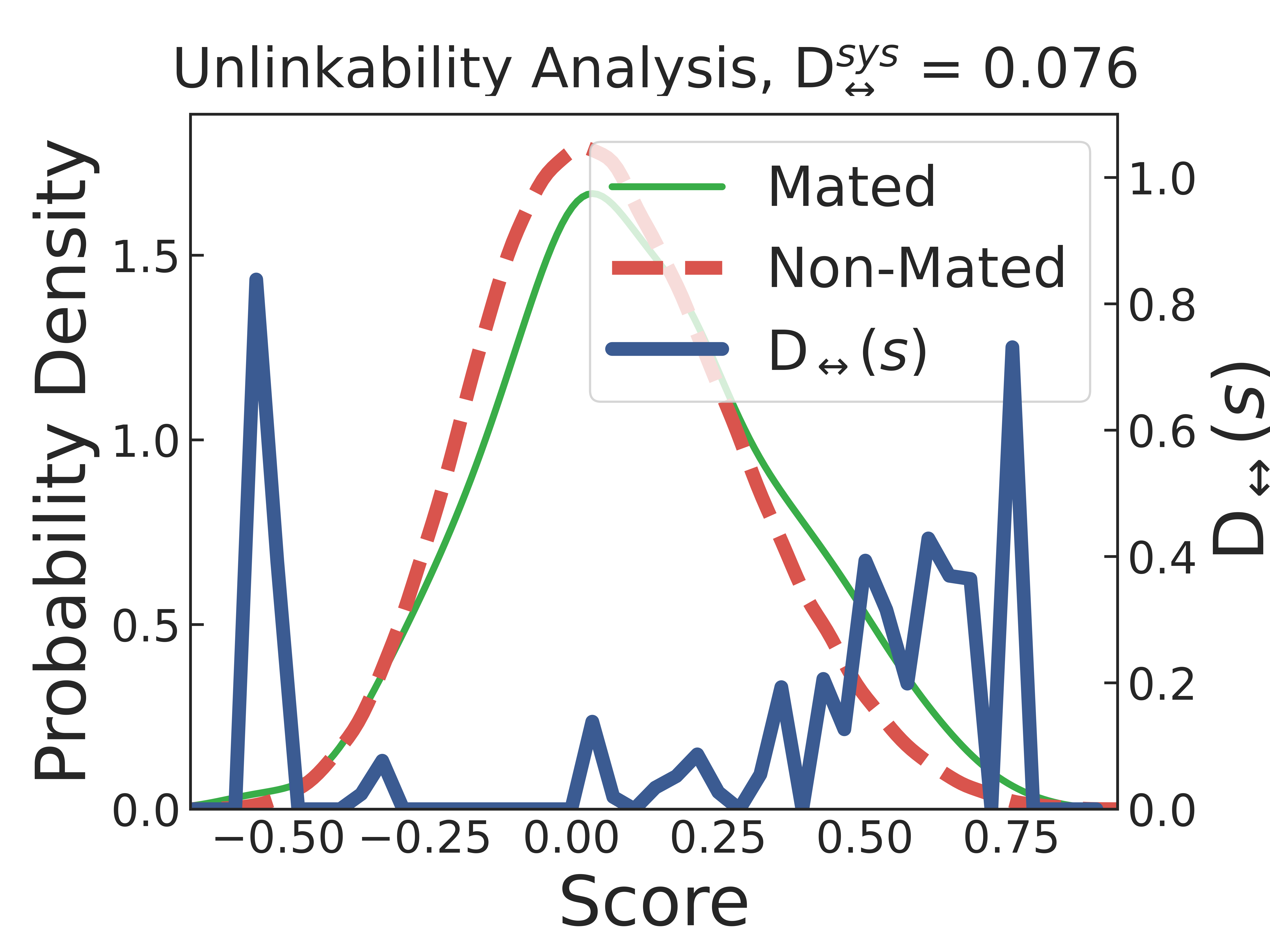}}
	\subfloat[\hspace{0.8em}\scriptsize{(d) PUT-WV}]{
		\includegraphics[width=0.45\linewidth]{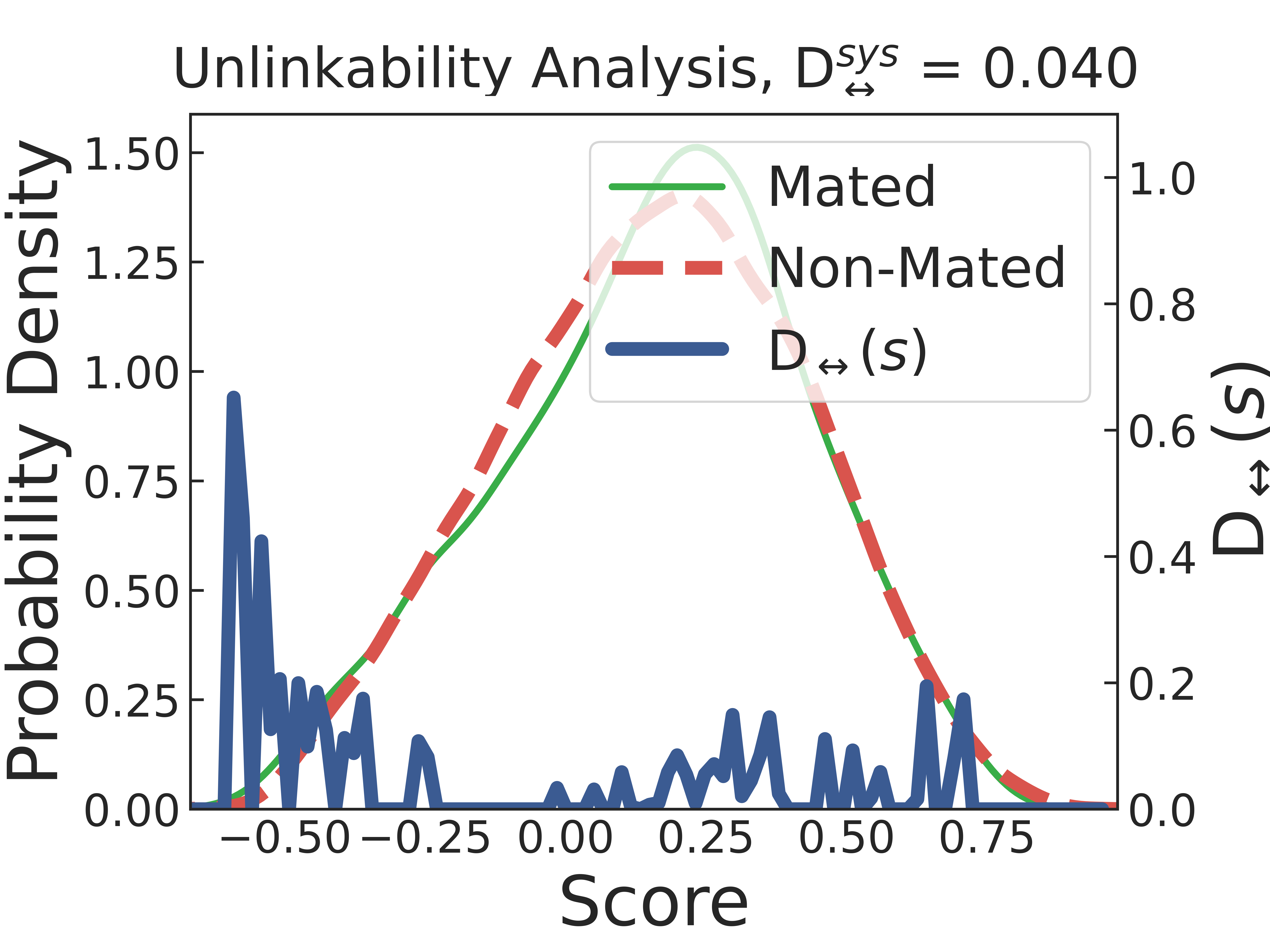}}
	\caption{Unlinkability analysis for ColorVein based cancelable identification systems on each dataset.}
	\label{fig11} 
\end{figure}
Unlinkability is a key attribute of cancelable biometrics, it requires that there be no linkage between protected templates generated using different tokens. This attribute ensures that users can generate protected templates that are unlinked in different color spaces when they enroll in different applications/databases. Thus, the adversary cannot launch an attack by analyzing the links between the user's protected templates across different applications.

We evaluated the unlinkability of the ColorVein-based cancelable vein recognition system using the evaluation framework proposed by Gomez-Barrero et al. \cite{gomez2017general}. Specifically, we compute mated and non-mated scores by setting different tokens for users to cross-match, and calculate the global linkability metric $D_\leftrightarrow^{sys}$. Fig. \ref{fig11} shows the distribution of mated and non-mated scores on each dataset. The results show that these two distributions highly overlap in most regions, indicating that the system shows well unlinkability. The ColorVein-based biometric system shows very low linkability, with $D_\leftrightarrow^{sys}$ close to 0. And only in cases when the mated scores were slightly higher than the non-mated scores did the system show slight linkability, which is reflected in the $D_\leftrightarrow{(s)}$ curve.

\subsection{Revocability}
\label{sec:revocability}
\begin{table}[]
	\renewcommand\arraystretch{0.5}
	\setlength\tabcolsep{5pt}
	\centering
	
	\caption{Decidability index ($d^\prime$) between the three distribution on each dataset. $d^\prime_{GI}$, $d^\prime_{GP}$ and $d^\prime_{IP}$ are the discriminability indices between genuine/impostor score distributions, genuine/pseudo-impostor score distributions and impostor/pseudo-impostor score distributions.}
	\label{tb8}
	\begin{tabular}{ccccc}
		\toprule
		Dataset & HKPU-FV &PUJ-DHV &PUT-PV & PUT-WV \\ \midrule
		$d^\prime_{GI}$       & 2.553   & 2.477     & 1.996&1.990 \\ \midrule
		$d^\prime_{GP}$       & 2.373    & 2.245     & 2.012 &2.073\\ \midrule
		$d^\prime_{IP}$       & 0.076    & 0.064      & 0.319 &0.443 \\ \bottomrule
	\end{tabular}
\end{table}

\begin{figure}[t]
	\captionsetup[subfloat]{labelsep=none,format=plain,labelformat=empty}
	\subfloat[\hspace{1.4em} \scriptsize{(a) HKPU-FV}]{
		\includegraphics[width=0.45\linewidth]{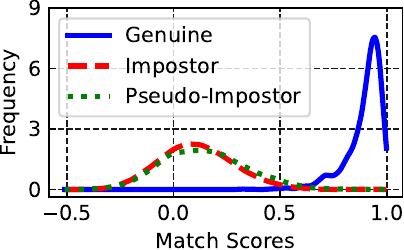}}
	\subfloat[\hspace{1.8em}\scriptsize{(b) PUT-PV}]{
		\includegraphics[width=0.45\linewidth]{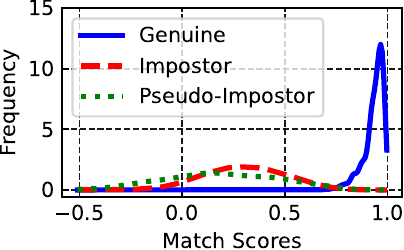}}
	\vspace{-10pt}
	\subfloat[\hspace{1.6em}\scriptsize{(c) PUJ-DHV}]{
		\includegraphics[width=0.45\linewidth]{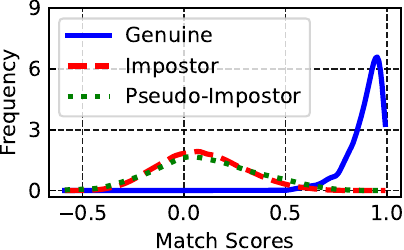}}
	\subfloat[\hspace{1.8em}\scriptsize{(d) PUT-WV}]{
		\includegraphics[width=0.45\linewidth]{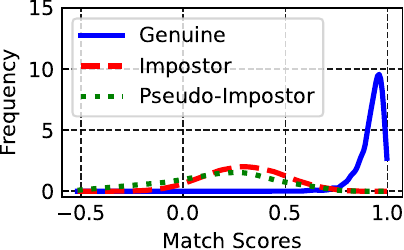}}
	\caption{The Genuine, Impostor and Pseudo-impostor match score distribution on each dataset.}
	\label{fig12} 
\end{figure}
In cancelable biometric systems, revocability is also a key attribute that refers to the fact that even if the original (stolen) and newly generated (regenerated) templates originate from the same vein instance, the two templates should be unrelated. This property can be verified by generating and analyzing pseudo-impostor distributions. Specifically, the cross-system user impostor distribution is referred to as the pseudo-impostor distribution. To empirically verify revocability, two conditions need to be satisfied (1) that the impostor and pseudo-impostor distributions are highly overlapping, and (2) that the genuine and pseudo-impostor distributions are unequivocally separable.

Fig. \ref{fig12} shows the score distributions of genuine, impostor and pseudo-impostor on each dataset. It can be clearly observed that there is a significant overlap between the score distributions of impostors and pseudo-impostors, and the score distributions of genuine and pseudo-impostors are clearly distinguishable. The degree of separability or overlap between such distributions can be assessed quantitatively by the $decidability$ $index$ $d^\prime$. Table \ref{tb8} lists the $d^\prime$ between the three distributions on each dataset. The$d^\prime$ between the distributions of impostors and pseudo-impostors is low in all datasets. There is good separability between the genuine and pseudo-impostor distributions. These results demonstrate that the ColorVein method fully satisfies revocability.

\subsection{Cancelable Biometric Attributes Comparison}
\label{sec:cbac}
As analyzed in Section \ref{sec:RPC}, among all types of vein biometrics, ColorVein achieves the highest recognition performance in terms of EER and it can secure the stolen token scenario. Next is biohashing, which although has good recognition performance for normal scenarios, but is not ideal for stolen scenarios. Bloom Filters on the contrary, barely perform the recognition task for normal scenarios, but can fully distinguish templates generated using stolen tokens. Block remapping and mesh warping are relatively low. For irreversibility, firstly Biohashing and bloom filters show a high level of irreversibility, helped by that they project vein features to a discrete binary space. And ColorVein gets almost the same irreversibility as them without information loss. Block remapping and mesh warping can strengthen the irreversibility by using smaller blocks or meshes.
Bloom filters, ColoVein and Biohashing all have high unlinkability and revocability, which indicates that these schemes can generate more diverse cancelable templates. Similarly, block remapping and mesh warping can use smaller blocks or meshes, increasing the diversity of generated templates.
In summary, ColorVein has a more comprehensive performance, and it is a cancelable biometric template generation solution specifically designed for vein biometrics. Moreover, from the perspective of vein pattern characteristics, ColorVein creatively uses color as the key to design cancelable templates, and enriches the feature information of vein images to improve the recognition performance.
\begin{figure}[t]
	\centering
	\includegraphics[scale=1,width=0.35\textwidth]{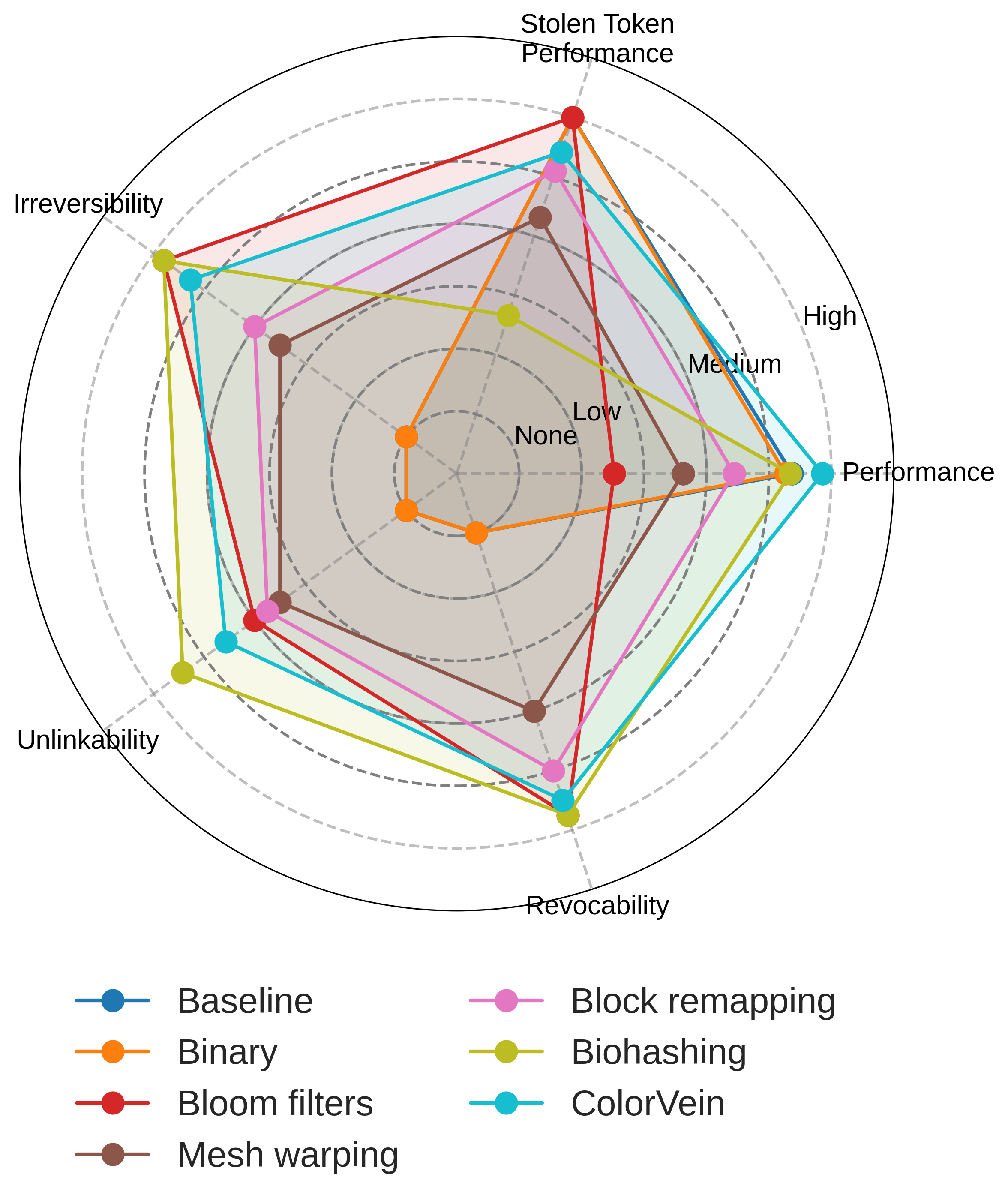}
	\caption{Comparison of Recognition Performance, Stolen Scenario Performance, Unlinkability, Irreversibility, and Revocability between ColorVein and representative cancelable schemes.}
	\label{fig13}
\end{figure}
\section{Conclusion}
\label{sec:conclusion}
This paper proposes an innovative cancelable vein biometric generation scheme: ColorVein, which is the first cancelable template generation scheme specifically designed for vein biometrics. By transforming static grayscale information into dynamically controllable color representations, ColorVein not only significantly enhances the information density of vein images but also provides a flexible mechanism for generating cancelable templates. Our scheme allows users/administrators to define a controllable pseudo-random color space for grayscale vein images by editing the location, number, and color of hint points, thereby generating protected cancelable templates. Furthermore, our proposed secure center loss further optimizes the training process of the protected feature extraction model, effectively increase the feature distance between legitimate enrolled users and potential impostors. Comprehensive evaluations of ColorVein on all vein biometrics (including finger, palm, dorsal hand, and wrist) demonstrate that we have confirmed the scheme's superior performance in terms of recognition performance, unlinkability, irreversibility, and revocability. Compared to existing state-of-the-art cancelable biometric schemes, ColorVein exhibits highly competitive performance. Overall, ColorVein provides a new perspective to cancelable vein biometric generation.

\bibliographystyle{IEEEtran}
\bibliography{ref}

\end{document}